\documentclass[]{interact}

\usepackage{epstopdf}
\usepackage{subfigure}

\usepackage{booktabs}
\usepackage{dsfont}
\usepackage{algorithm}
\usepackage{algorithmicx}
\usepackage{algpseudocode}
\usepackage{algcompatible}
\usepackage{multirow}
\usepackage{url}
\usepackage[normalem]{ulem}

\usepackage{natbib}
\usepackage{apalike}
\usepackage{appendix}

\theoremstyle{plain}

\theoremstyle{definition}

\theoremstyle{remark}

\usepackage{xcolor}
\newcommand{\change}[1]{{\color{black}{#1}}}

\begin{document}

\articletype{ARTICLE TEMPLATE}

\title{Guided Navigation from Multiple Viewpoints using Qualitative Spatial Reasoning}

\author{
\name{D.~H. Perico\textsuperscript{a}\thanks{CONTACT D.~H. Perico. Email: dperico@fei.edu.br. Department of Computer Science, Centro Universitáio FEI, Avenida Humberto de Alencar Castelo Branco, 3972, S\~ao Bernardo do Campo, SP, Brazil}, P.~E. Santos\textsuperscript{a,b} and R.~A.~C. Bianchi\textsuperscript{a}}
\affil{\textsuperscript{a}Centro Universitário FEI, S\~ao Bernardo do Campo, Brazil; \\ \textsuperscript{b}Flinders University, Adelaide, Australia}
}

\maketitle

\begin{abstract} 
Navigation is an essential ability for mobile agents to be completely autonomous and able to perform complex actions. However, the problem of navigation for agents with limited (or no) perception of the world, or devoid of a fully defined motion model, has received little attention from research in AI and Robotics. One way to tackle this problem is to use guided navigation, in which other autonomous agents, endowed with perception, can combine their distinct viewpoints to infer the localisation and the appropriate commands to guide a sensory deprived agent through a particular path. Due to the limited knowledge about the physical and perceptual characteristics of the guided agent, this task should be conducted on a level of abstraction allowing the use of a generic motion model, and high-level commands, that can be applied by any type of autonomous agents, including humans. The main task considered in this work is, given a group of autonomous agents perceiving their common environment with their independent, egocentric and local vision sensors, the development and evaluation of algorithms capable of producing a set of high-level commands (involving qualitative directions: e.g. {\em move left}, {\em go straight ahead}) capable of guiding a \change{{\em sensory deprived}} robot to a goal location. In order to accomplish this, the present paper assumes relations from the qualitative spatial reasoning formalism called StarVars, whose inference method is also used to build a model of the domain. This paper presents two qualitative-probabilistic algorithms for guided navigation using a particle filter and qualitative spatial relations. In the first algorithm, the particle filter is run upon a qualitative representation of the domain; whereas, the second algorithm transforms the numerical output of a standard particle filter into qualitative relations to guide a sensory deprived robot. The proposed methods were evaluated with experiments carried out on a 2D humanoid robot simulator. A proof of concept executing the algorithms on a group of real humanoid robots is also presented. The results obtained demonstrate the success of the guided navigation models proposed in this work.
\end{abstract}


\section{Introduction}

Navigation is an essential ability of any autonomous mobile agent. This is a complex task that involves several cognitive competences, such as the elaboration and interpretation of maps, localisation and the decision of which action to take in order to maintain a desired path \citep{COLOMBO2017605}. Although these issues have been tackled extensively in robotics \citep{desouza02,KRUSE20131726}, spatial cognition \citep{Epstein17}, and at the intersection of both fields \citep{ekstrom18}, little attention has been given to the investigation of algorithmic methods that present a solution to cases in which the agent has to navigate with limited perception, no sensing apparatus or without a defined motion model. From the spatial cognition standpoint, this issue was briefly considered within the description or depiction of maps and route directions \citep{tversky99,Tversky2014}. \change{Inspired by such work, the present paper considers the problem of autonomous guided navigation. Guided navigation is the task} in which a group of autonomous agents (observers) use a combination of the information obtained from their multiple (distinct and egocentric) viewpoints to infer the pose, the route and the actions that a guided, sensory deprived, agent must execute to achieve a goal destination. It is desirable that this navigation system should be able to guide any type of mobile autonomous agent (human or robotic), as long as the agent is capable to follow commands in terms of direction relations. To this end, the present paper assumes commands defined upon a set of qualitative directions (e.g. {\em move left}, {\em go straight ahead}) that could be easily interpreted by a human agent, but that should also be \change{translatable} to low-level robot actions. The main task considered in this work is, given a group of autonomous agents perceiving their common environment with independent, egocentric and local vision sensors, the development and evaluation of algorithms capable of producing a set of high-level commands capable of guiding a {\em sensory deprived} robot to a goal location. The success rate for this task is, then, measured as the number of trials in which the sensory deprived agent is able to reach the goal location by using only the information provided by the observers.

This work falls within the cognitive robotics field, whose goal is the investigation of ``the knowledge representation and reasoning problems faced by an autonomous robot (or agent) in a dynamic
and incompletely known world" \citep{levesque08}. Although the present paper is mainly concerned with algorithmic aspects of guided navigation considering groups of robots, we believe that this work could open avenues of research toward the development of navigation strategies for robots and humans considering a human-like description of maps, routes and self-localisation \citep{tversky99,Tversky2014,Wolter2008}. The cognitive adequacy of this line of research has been considered in \citep{LueckeEtAL2011}, where a set of qualitative relations (similar to those used in this paper) was designed to emulate the mental conceptualisations of wayfinding and route elements proposed in \citep{KLIPPEL2005311}.

We can envision a number of application scenarios for the investigation described in this paper. \change{Considering a mobile robot, the need to receive commands from other agents can arise due to \change{possible} failure of its perceptual system while working on a collaborative task (e.g. search and rescue)}. Another domain where guided navigation comes into play is the collaboration of multiple agents toward the execution of a common task. In this case, the robot that is the closest to achieve a goal may momentarily have his perception impaired and be in need of guidance (this is a common situation in robot\footnote{(or human!)} soccer). We can also consider the application of these ideas considering humans as guided agents where, for instance, sensors located in the environment could provide wayfinding assistance to people with partial or total visual impairment, or in military domains where surveillance drones could provide essential routes to platoons engaged in conflict situations. These application scenarios provide the motivation to develop this research, however the actual deployment of the results presented here in real-world domains is outside the scope of this paper.

In this paper, the guided agent receives commands from the observer agents in terms of qualitative spatial relations, such as {\em move right}, {\em move left}, {\em move forward} and {\em move backward}. These commands are modelled using the qualitative spatial reasoning formalism called $StarVars$ \citep{Lee13starvars} (Section \ref{sv}), that is a spatial calculus defined upon relative directions that can be effectively used to integrate observer-relative information. $StarVars$ is used alongside a particle filter to provide guided navigation within a world model constructed using the available multiple viewpoints of observers in a domain. Particle filter (PF) \citep{Thrun:2005} (Section \ref{pf}) is a traditional filtering technique that works in two stages: update and prediction. In this paper two PF algorithms are introduced: in the first, the update stage of the filter was modified to receive, instead of the usual numerical data, a qualitative world model that is constructed using the $StarVars$ calculus (Section \ref{qpfa}); in the second, the particle filter runs upon the relative locations of the various objects in the domain, and the
$StarVars$ representation is used only to transform the outputs of the filter into qualitative relations in terms of the agents’ egocentric reference frame (Section \ref{npfgnqi}). Results (Section \ref{results}) show that both algorithms provide effective guidance to a sensory deprived agent, where the second method proposed shows a more robust behaviour.

\section{Background}

This section presents a summary of the main techniques used in this work, namely Particle Filter and StarVars Calculus.

\subsection{Particle Filter}\label{pf}

Particle Filter \citep{Thrun:2005} is the generic name for some probabilistic techniques whose purpose is to recursively compute, at each time instant~$t$, the posterior probability $bel$ of a set $S_t$ of $N$ particles randomly distributed and composed of a weight $w$, assigned according to the particle's importance. Informally, a particle filter algorithm, applied to a robot's pose, generates a number of hypotheses of the possible locations and orientations of the robot and tests them against sensor information. Particles that agree with the observations receive a higher weight than particles that do not agree (the latter end up being ignored as the inference progresses). 

The basic Particle Filter method is shown in Algorithm \ref{algo:particle_filter}, where $S_{t-1}$ represents the particle set at $t-1$, $u_t$ represents an input that influences the transition model and $z_t$ represents a measurement. Each particle in $S_{t-1}$ represents a possible state of the system (i.e. the robot's pose, in the present context). In Line $3$ of Algorithm \ref{algo:particle_filter}, a state $x_t^i$ is created based on the particle $x_{t-1}^i$ and the action $u_t$. The set obtained after $N$ iterations of this line represents the predicted belief $ \overline{bel}(x_t)$. In turn, at Line $4$, the $w_t^i$ importance factor of each particle (its weight) is calculated based on the $z_t$ measurement. In fact, $w_t^i$ is the probability of obtaining the measurement $z_t$ with the particle $x_t^i$. The set of particles (with their respective weights) represents the posterior belief $bel(x_t)$. Lines 7 to 10 represent the particle resampling (or importance sampling), where the algorithm draws particles from the temporary $\overline{S_t}$ and inserts them into the new set ${S_t}$. The probability of a particle $x_t^i$ being sampled is given according to its importance factor $w_t^i$. Prior to resampling, the particles are distributed according to $\overline{bel}$. After that, the distribution becomes the posterior belief $bel(x_t)$.

\begin{algorithm}[t]
\caption{Particle Filter}\label{algo:particle_filter}
\begin{algorithmic}[1]
\REQUIRE \(S_{t-1}, u_t, z_t\)
\vspace{0.2cm}
\STATE $\overline{S_t}$ = $S_t$ = $\emptyset$ 
\FOR{$i = 0$ to {$N$}}
\STATE Sample $x_t^i$ $\sim$ $p(x_t|u_t, x_{t-1}^i)$ 
\STATE $w_t^i = p(z_t | x_t^i)$ 
\STATE $\overline{S_t} = \overline{S_t}$ $\cup$ $<x_t^i, w_t^i>$ 
 \ENDFOR
 \FOR{$i = 0$ to {$N$}}
\STATE draw $m$ with probability $\propto$ $w_t^m$ 
\STATE push $x_t^m$ into $S_t$
 \ENDFOR
\STATE RETURN \(S_t\)

\end{algorithmic}
\end{algorithm}

\subsection{The $StarVars$ calculus}\label{sv}

The Star Calculus with Variable interpretation of orientation  - $StarVars_m$ \citep{Lee13starvars} is a qualitative spatial calculus defined upon relative directions based on the absolute direction calculus ${Star}_m$. The ${Star}_m$ calculus \citep{STARRenz04} is a formalisation of qualitative directions between pairs of points (with respect to a given reference direction). It is defined by $n$ lines intersecting at each point, thus defining $2n$ sectors per point, where $m = 2n$ is the calculus granularity. This calculus is part of the Qualitative Spatial Reasoning (QSR) field of research \citep{QSRKRHandbook07,ligozat2013}. QSR is a subfield of knowledge representation in artificial intelligence that attempts the formalisation of spatial knowledge based on primitive relations defined over elementary spatial entities.

\begin{figure}[t!]
\centering
\centering
\includegraphics[scale=0.5]{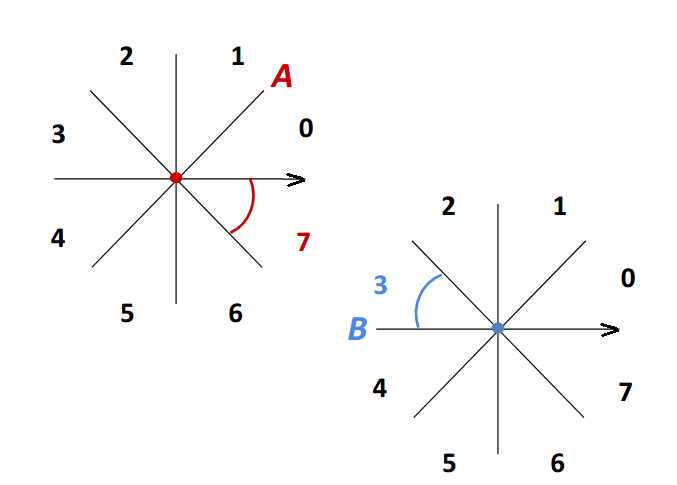}
\caption{${Star}_8$ relations: $A (7) B \wedge B (3) A$, as introduced in \cite{JaeHeeLee}.}
\label{fig:star}
\end{figure} 

A qualitative direction relation in ${Star}_m$, $p(R)q$ is the sector $R$ of $p$ in which $q$ is located \citep{Lee13starvars}.  Figure \ref{fig:star} shows an example of ${Star}_8$ relations (or constraints) between two points $A$ and $B$: $A(7)B$ and $B(3)A$, where $A(7)B$ represents that point $B$ is in the sector (7) of $A$ and point point $A$ is in the sector (3) of $B$. 

The notation $ [a, b [$ is used in  \cite{JaeHeeLee} to represent the angular sector delimited by the lines $ a $ and $ b $. Reasoning with ${Star}_m$ can be executed using composition tables \citep{STARRenz04}. However, \cite{JaeHeeLee} argues that this can be more efficiently accomplished using Linear Programming (LP). Linear programming \citep{rlinear} is a widely used method for solving optimisation problems. A linear program is composed of an objective linear function and a system of linear equations and inequalities called constraints, whose solutions are obtained by minimising (or maximising) the objective function while respecting the constraints. A ${Star}_m$ set of relations can be described as a linear program, whereby the system of inequalities is constructed based on vector algebra in the following way:  given two vectors $\vec{v}, \vec{u} \in \mathds{R}^2 $, each represented as a column of a $ 2 \times 2 $ matrix; $\vec{u}$ is on the left (right) of $\vec{v}$ if and only if the determinant of these two vectors is greater (less) than zero, $det(\vec{v}, \vec{u})> 0 $ ($det(\vec{v}, \vec{u}) < 0 $); $\vec{u} $ is parallel to $ \vec{v} $ if and only if $ det (\vec{v}, \vec{u}) =  0$. 

Let $\vec{v} = (x_2, y_2) - (x_1, y_1) \in \mathds {R}^2$ be a vector with start point $A = (x_1, y_1)$ and end point $B = (x_2, y_2)$ and a unit vector $u(A, s)$ with the same orientation as the line $ s $ from $A$; that is, $u(A, s) = (cos (\theta_1 + s \eta), sen(\theta_1 + s \eta))$, where $\theta_1$ is the orientation of $A$ and $\eta = \frac{360^o} {m}$. The relation $A (s) B$ is satisfied if and only if $\vec{v}$ is on the left or parallel to $u(A, c)$ and $\vec{v}$ is on the right of $u(A, d)$, considering the convex relationship $A [c, d[B$. So, $det(u(A, c), \vec{v}) \geqslant 0$ and $det(u(A, c), \vec{v}) <0 $, which is equivalent to the following system of inequalities\footnote{As linear programming is only defined upon non-strict inequalities, the second inequality in the system \eqref{non_strict} must be adapted to be less than or equal to an infinitesimal negative value $\epsilon$ (as shown in Eq. \ref{non_strict2}).}:
 \begin{align}
 \begin{cases}\label{non_strict}
 -sin(c\eta)x_1 + sin(c\eta)x_2 + cos(c\eta)y_1 - cos(c\eta)y_2 \leqslant 0, \\
 sin(d\eta)x_1 - sin(d\eta)x_2 - cos(d\eta)y_1 + cos(d\eta)y_2 < 0.
 \end{cases}
 \end{align}

${StarVars}_m$  is defined on $\mathds{R}^2 \times \Theta_m $, where $m$ represents the granularity ($m \in \mathds{N}$ and $ m \geqslant 2 $) and $ \Theta_m = \{0 \eta, 1 \eta, 2 \eta, ..., (m-1) \eta \} $, the set of orientations in the domain with $ \eta = 360^o / m $. A ${StarVars}_m $ spatial object is composed of $ (x, y, \theta) $, where $ (x, y) \in \mathds {R}^2 $ determines the position and $ \theta \in \Theta_m $ the orientation \citep{JaeHeeLee}.

${StarVars}_m $ relations have the same meanings as ${Star}$ relations. However, the former relations are interpreted according to the individual orientations of each spatial object, where the $ 0 $ line is always aligned with the $\theta$ orientation. Thus, reasoning with ${StarVars}_m $ can be done using linear programming by including the orientation of the spatial object in the system of inequalities \eqref{non_strict}, resulting in the system of inequalities \eqref{non_strict2} \citep{Lee13starvars}, considering the convex relationship $[c, d [$ between two spatial objects.
 \begin{align}
 \begin{cases}\label{non_strict2}
 -sin(c\eta + \theta_1)x_1 + sin(c\eta + \theta_1)x_2 + cos(c\eta + \theta_1)y_1 -\\ \hspace{4cm}  cos(c\eta + \theta_1)y_2 \leqslant 0, \\
 sin(d\eta + \theta_1)x_1 - sin(d\eta + \theta_1)x_2 - cos(d\eta + \theta_1)y_1+ \\ \hspace{4cm} cos(d\eta + \theta_1)y_2 \leqslant \epsilon.
 \end{cases}
 \end{align}
 
Using the granularity $m$ as input, the number of spatial entities $n$ and the set of direction relations between all entities (referred as $\phi$, where $\phi = \bigwedge_{i \neq j} v_i R_{ij} v_j$, where $i, j = 1, ..., n, i \neq j$), the solution of the system of inequalities \eqref{non_strict2} give the orientations of all spatial entities in the domain. Linear programming then returns a world model $\psi = \bigwedge_{i \in \{1, ..., n \}} (x_i, y_i, \theta_{i})$ as soon as the first set of orientations solving the inequality system is found. Otherwise, LP continues the search and returns $Fail$ if no such set is found. Thus, $StarVars_m $ returns a complete world model, with $x$, $y$, and $\theta$ from each oriented spatial entity and $x$ and $y$ from each non-oriented reference point, if a world model exists. In this work, the simplex method \citep{Dantzig:1990} was used to solve the linear programs.

For example, let's assume a domain with three spatial entities, two of them ($A$ and $B$) have orientations and one ($C$) is a non-oriented point. And, let the set of qualitative direction relations between these entities be: $\{A(1)B \wedge B(0)A \wedge A(0)C \wedge B(2)C\}$, for $m = 8$. In this case, it is possible to define a system of $8$ inequalities, where the variables are the orientations $ \theta_A $ and $ \theta_B $ of points $ A $ and $ B $. This system is obtained from the system of inequalities \eqref{non_strict2} by replacing $c$ with the relations $\{(1), (0), (0), (2)\}$ (between $A, B$ and $C$, as above) and $d$ by $c + 1$. The resulting linear program is used to check whether the system can be satisfied for different values of $ \theta_A $ and $ \theta_B $. If consistent values are found, a world model with coordinates and orientations for all domain elements is returned. In this example, a possible world model (assuming $ \epsilon = -1 $) would be: $x_A = 0.0$, $y_A = 0.0$ and $\theta_A = 0^o$; $x_B = 1.0$, $y_B = 1.0$ and $\theta_B = 225^o$; $x_C = 2.0$ and $y_C = 0.0$.

 \label{rev3} \change{ StarVars reasoning starts by checking the consistency of the inputs. If consistency is verified, the system returns at least one world model. In case the system does not return any model, it can be said that the input is inconsistent. As each observer agent has an orientation given by $\theta \in \Theta_m$, where $\Theta_m = \{0\eta, 1\eta, 2\eta, ..., (m-1)\eta\}$ and $\eta = 360^o/m$, if any agent is in a different orientation than that represented by $\Theta_m$, the model fails. $StarVars$ also fails to provide a model if errors in the object detection occur, leading to the incorrect description of the relations holding between spatial entities. Moreover, the model returned by $StarVars$ may not be unique (but all resulting models are consistent with respect to the observations); thus, as we shall see further in this paper, the first model generated is chosen at the beginning of each $StarVars$ iteration.}

\subsection{Agent Model in $StarVars$}\label{agent}

\begin{figure}[t!]
\centering
\includegraphics[scale=0.6]{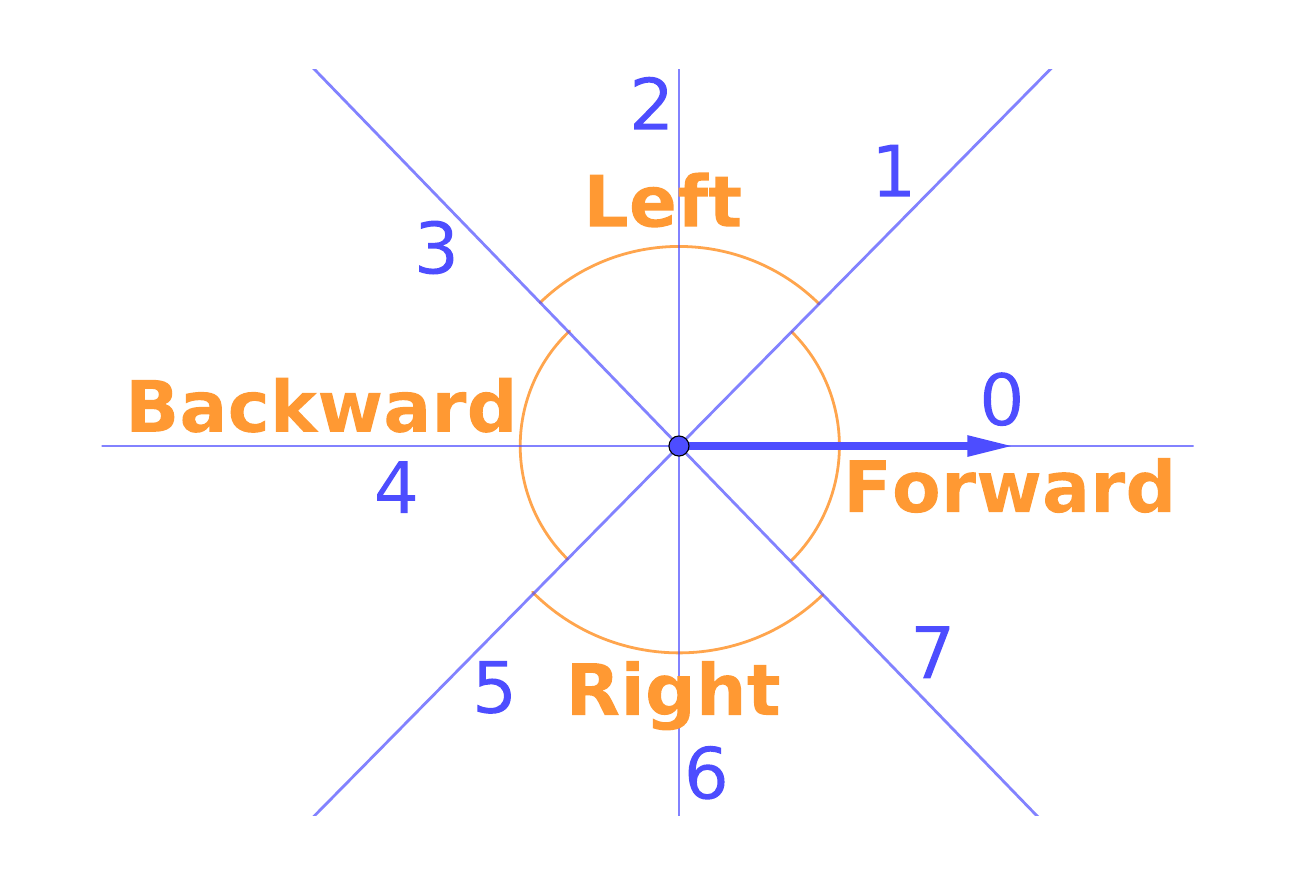}
\caption{Guided agent directions for $m$ = 8.}
\label{fig:projetivas}
\end{figure}

The angular sectors of $StarVars_m$ are combined to represent qualitative relative directions in terms of the agents' egocentric reference frame, according to Equations \eqref{eq:esq} to \eqref{eq:frente} (cf. Figure \ref{fig:projetivas}).

\begin{align}
   Le{f}t &= [\frac{m}{8}, \frac{3m}{8}[,& \label{eq:esq}\\
   Right &= [\frac{3m}{8}, \frac{5m}{8}[,&\label{eq:atras}\\
  Forward &= [\frac{5m}{8}, \frac{7m}{8}[,&\label{eq:dir}\\
  Backward &= [\frac{7m}{8}, \frac{m}{8}[.&\label{eq:frente}
\end{align}

\sloppy
The guided agent follows simple spatial expressions (actions) such as $Stop$, $Turn Right$, $Turn Left$, $Move Forward$ and $Move Backward$ (defined in terms of $StarVars_8$) in order to reach a goal region. Let $\delta_t$ represent a Gaussian noise with zero mean and variance $\sigma$ ($\delta_t \sim \mathcal{N} (0, \sigma)$), these actions are then defined as follows:

\begin{itemize}
\item $Stop$: no movement is executed;
\item $Move Forward$: the agent moves in a straight line with respect to its orientation at any speed;
\item $Turn Right$: the agent rotates $-90^o + \delta_t$ and applies a $Move Forward$ action;
\item $Turn Left$: the agent rotates $+90^o + \delta_t$ and applies a $Move Forward$ action;
\item $Move Backward$: the agent rotates $180^o + \delta_t$ and applies a $Move Forward$ action;
\end{itemize}

The {$Move forward$} action only ends when the guided agent is instructed to stop. 

The guided agent has no information of its pose and does not have a direct perception of the domain, apart from receiving commands as the spatial expressions listed above. The other agents (called {\em observer agents}), on the other hand, have total knowledge of their poses and perceive their environment by means of cameras located in their ``heads" (assuming humanoid robots). I.e. the observer agents can only share their local views of the environment, no global viewpoint is available. This work assumes that there could be any finite number of observer agents in the domain, but only one guided agent is allowed. Regardless of the number of observer agents present, the commands communicated to the guided agent are passed  only by one of the observer agents, called {\em coordinator}. It is the coordinator agent that makes the necessary inferences given the perspectives and orientations of all other observer agents. The $Stop$ command is the only exception, as it can be issued by any observer agent at any time. Any agent can play the role of coordinator, but once assigned, this role should not change until the end of a navigation task.

\section{Guided Robot Navigation}\label{algo}

This section describes a general algorithm for guided navigation and two distinct instantiations of it: the {\em Qualitative Particle Filter} (QPF) algorithm (Section \ref{qpfa}) and the {\em Particle Filter with Qualitative Commands} (PFQC) algorithm (Section \ref{npfgnqi}).  The former uses qualitative data to feed a particle filter that is responsible for taking the guided agent toward a goal location. In this model, both motion prediction and update are performed based on the qualitative regions defined by the intersection of the $StarVars_m$ sectors defined by the location of all agents in the domain. The latter uses numerical direction and orientation data to feed a conventional particle filter and, with these particles, to infer commands which are then transformed into qualitative relations for leading the guided agent to the goal.  In this work a guided navigation task is considered successful if the sensory deprived agent was able to reach the goal location by using uniquely the information provided by the other observers.

Videos with running examples of these algorithms are available in the following sites: 
\begin{itemize}
    \item For simulation examples: \url{https://youtu.be/kGyisRRThjY}
    \item For examples with real robots: \url{https://youtu.be/aoiU1JjJg1M}
\end{itemize}

\subsection{General Guided Navigation Algorithm}\label{general}

In the general guiding agent algorithm (Algorithm \ref{algo:geral}) the set of all spatial entities ($e_j$) is represented by $SE$ and consists of all observer agents ($o_i$), a coordinator agent ($o_c$) and the guided agent ($g$) (that does not have a direct perception of its environment). Informally, the general algorithm works as follows: the coordinator agent ($o_c$) first collects all the locations and observations of all observer agents $o$, $o = \{o_i\}\cup \{o_c$\}, in the domain; $o_c$ then uses this information to create a global representation (a map) with the poses of all observers and objects perceived by each observer agent $o$; this map is used to generate particles, hypotheses about all the possible poses of $g$ and every observer $o$. The particles with the highest probability values guide the choice of actions (commands) that are sent to $g$; as $g$ moves, the other agents' perceptions are updated, as well as the particles' probability values assigned to the domain entities (as presented in Section \ref{agent}). The algorithms QPF and PFQC, described below, are instances of this process, differing according to the way the map is constructed and how the particles are distributed and evaluated.

The set of relative directions (as perceived by all observer agents) is represented by $Z_c$, where $Z_c = \bigwedge_{i\neq j} < o_i, z_{ij}, e_j, \theta_i>$ (where $i$ runs over the set of observer agents and $j$ over the set of all domain elements), $z_{ij}$ is the $StarVars$ relation holding between $o_i$ and $e_j$ and $\theta_i$ is the orientation of observer $o_i$.

\begin{algorithm}[!t]
\caption{General guiding agent algorithm}\label{algo:geral}
\begin{algorithmic}[1]
\REQUIRE {$SE$ = Set of all spatial entities $e_j$, consisting of observing agents ($o_i$), guided agent ($g$), and goal;  $n_o$ = the total number of observer agents; $n$ = the total number of spatial entities; $m$ = the granularity of $StarVars$}

\STATE	$Z_c = \emptyset$ 
	\FOR{$o_i (i \in \{1,..,n_o\})$, $e_j (j \in \{1,...,n\})$} \label{l1a}
			\IF {$o_i \neq e_j$} 
			\Statex \change{\% {\em Each observer sends their observations to the coordinator agent}}
				\STATE $z_{ij}$ = \texttt{getDirection($o_i,e_j$)} \label{zij}
				\STATE $o_i$ sends $ <o_i, z_{ij}, e_j, \theta_i> $ to the coordinator agent $o_c$
				\STATE $Z_c = Z_c$ $\cup$ $<o_i, z_{ij}, e_j, \theta_i>$
			\ENDIF
	\ENDFOR \label{l1b}
\Statex \change{\% \em Guided agent $g$ has information about his own orientation}		
	\IF {g's orientation ($\theta_g$) is known}\label{11}
		\STATE $g$ sends $ <g, \theta_g> $ to the coordinator agent $o_c$
		\STATE $Z_c = Z_c$ $\cup$ $<g, \theta_g>$ 
    \ENDIF \label{14}
\Statex \change{\% \em With all the observations collected, mapping and particle release is executed}
\STATE	$\psi$ = \texttt{mapping}$_{c}(Z_c)$ \label{mapping}
\STATE	$S$ = \texttt{releaseParticles}$_{c}(\psi)$ \label{release} 
\Statex \change{\% \em Main Loop: prediction and update}
	\WHILE{$g$ does not reach the goal}\label{17}
	\STATE	$action$ = \texttt{chooseAction}$_{c}^{g}()$ \change{\% The most probable action is chosen}\label{18}
	\STATE	$\overline{S_{g}}$ = \texttt{predict}$_{c}(action, S_{g})$ \label{21}
	\STATE	\texttt{submit}$_{c}^{g}(action)$ \label{sub}
		\WHILE{$g$ is not in stopping condition}\label{23}
		\STATE	Move guided agent $g$ according to $action$ 
		    \Statex \hspace{0.3in}\change{\% \em Observers update their perceptions} 
			\FOR{$o_i, i \in \{1,..,n_o\}$}
			\STATE	$z_{ig}$ = \texttt{getDirection($o_i,g$)} \label{26} 	
			\ENDFOR
		\ENDWHILE \label{ew1}	
		
\STATE		Guided agent $g$ stops moving 
\Statex \hspace{0.15in}\change{\% \em Coordinator collects the observations} 
		\FOR{$o_i, i \in \{1,..,n_o\}$}\label{28}
			\STATE $o_i $ sends $ <o_i, z_{ig}, g, \theta_i> $ to the coordinator agent $ c $ 
			\STATE replace $ <o_i, z_{ig}', g, \theta_i>$ for $ <o_i, z_{ig}, g, \theta_i>$ em $ Z_c$ 
		\ENDFOR
		
		\IF {the guided robot's orientation ($\theta_g$) is given } \label{32}
		\STATE	$g $ sends $ <g, \theta_g> $ to the coordinator agent $ o_c $
		\ENDIF  \label{35}
		\Statex \change{\% \em The particles are updated and the prediction-update loop starts again} 
		\STATE $S_{g}$ = \texttt{update}$_{c}(\overline{S_{g}}, Z_c)$ \label{31}
\ENDWHILE
\end{algorithmic}
\end{algorithm}

All guided navigation inferences and messages are executed by the coordinator agent $o_c$, so all functions attributed to it have a subscript index $c$. Some of these functions directly influence the guided agent, so they get a superscript $g$.

Informally, Algorithm \ref{algo:geral} can be described as follows:

\change{\subsubsection*{\% Each observer sends their observations to the coordinator agent}}

The loop between Lines \ref{l1a} and \ref{l1b} of Algorithm \ref{algo:geral} represents the procedure for collecting the directions (operated by the function $\texttt{getDirection}$) of each spatial entity $e_j$ as perceived by each observer agent $o_i$ ($z_{ij}$ in Line \ref{zij}). The case where the guided agent has information about its orientation is considered in Lines \ref{11}--\ref{14}.

\change{\subsubsection*{\% With all the observations collected, mapping and particle release is executed}}

The $ \texttt{mapping}_{c}$ function (Line \ref{mapping}) is responsible for creating the internal representation ($\psi$) of the location of all spatial entities in the domain, that is, a map of all agents and the goal. \change{In this work, this is accomplished only one time per execution, thus it is outside of the main loop. Future work will consider a more dynamic environment, where the observers and other domain objects can move, then this function should be included in the loop.} The $ \texttt{releaseParticles}_{c}$ function (Line $\ref{release}$) uses the $\psi$ world model to generate the set of particles $S = \{S_{o_{i = \{1, .., n_o\}}} \cup S_{g} \cup S_{goal} \}$, containing one particle for each observer agent (set $S_{o_{i = \{1, .., n_o \}}}$), several particles for the guided agent (set $S_{g}$), of which the uncertainty in pose is larger, and one or more particles for the goal (set $S_{goal}$). Each particle at location $(x,y)$ represents a pose and an importance factor for each agent: $\langle\langle x, y, \theta \rangle, w \rangle$, whereas the goal particles are defined only by their 2D position: $\langle x, y \rangle$.

\change{\subsubsection*{\% Main Loop: prediction and update}

Once the mapping is executed and the particles are released, the main loop starts (Line \ref{17}), in which the sensory deprived agent is guided using the perceptions of the other agents, and the actions selected accordingly.}

The $\texttt{chooseAction}_{c}^{g}$ function is in charge of  performing the guided agent motion planning, returning an $action$, or command, in terms of a high-level, qualitative, relation (as discussed in Section \ref{agent}); $\texttt{chooseAction}_{c}^{g}$ (Line \ref{18}) compares the relative position of the goal with respect to each particle representing the agent's $g$ pose. \change{\label{rev3.10} As there are $N$ particles, the action selected is the one with the highest probability (i.e., the most frequent, given the goal and the guided-agent particles).} Therefore, the decision about the guided agent actions is made as it moves through the domain. This $action$ is used by the coordinator to predict the pose of the guided agent (Line \ref{21}). At the beginning of each cycle of prediction and update, a new action choice is made by the coordinator agent.

\change{\subsubsection*{\% Observers update their perceptions}}

As the guided agent ($g$) moves, the observers constantly update the $g$'s perceived relative direction (Lines \ref{23} to \ref{ew1}) using the $\texttt{getDirection}$ function until a stopping criterion is met. Then, new directions $z_{ig}$ are sent to the coordinator agent, which updates the set $Z_c$ (Lines $\ref{28}$ to $\ref{31}$). \change{ A distinct stopping criterion is defined for each of the instances of Algorithm \ref{algo:geral} investigated in this work: QPF and PFQC, as we shall see further in this paper.}

If it is the case that the guided agent has knowledge about his own orientation, this information is also submitted to the coordinator, as represented in Lines \ref{32} to \ref{35} \change{(it is worth pointing out that this is done in the main loop here)}.

\change{\subsubsection*{\% The particles are updated and the prediction-update loop starts again}

The particles are updated with the new observations (in Line \ref{31}) and a new iteration of the main loop starts with the selection of another action on Line \ref{18}.}

The remainder of this section is dedicated to the introduction of two variants of the general model presented in Algorithm \ref{algo:geral}. In the algorithm presented in Section \ref{qpfa} (QPF) the input data is qualitative, whereas the algorithm presented in Section \ref{npfgnqi} (PFQC) the input is numeric. Thus, as we shall see, the $\texttt{getDirection}$, $\texttt{mapping}_{c}$, $\texttt{prediction}_{c}$ and $\texttt{update}_{c}$ functions are defined in distinct ways. In spite of that, both methods use qualitative commands to guide the sensory deprived agent to the goal. This makes the procedure for selecting the guided agent actions, $\texttt{chooseAction}_{c}^{g}$, the same in both methods.

 As mentioned above, there is no pose associated to the goal position, as it is solely defined by its 2D coordinate. Assuming a pose-independent goal is a simplifying constraint imposed at the beginning of this research. Relaxing this assumption should lead to fewer models being generated and to a longer execution time, but could extend the applicability of the methods proposed in this work to domains where (for instance) the goal completion involves the manipulation of objects, or to domains where a privileged view position is part of the goal. Relaxing this assumption is a task for future research.

\subsection{Qualitative Particle Filter (QPF) Algorithm}\label{qpfa}

\begin{algorithm}[t]
\caption{$\texttt{Mapping}_{c}$ for QPF}\label{a2}
\begin{algorithmic}[1]
\REQUIRE {\(Z_c, n_o, n \)}
\STATE $A,b$ = \texttt{buildInequalities}$(Z_c, n, n_o)$ 
\STATE $\psi$ = \texttt{searchValidModel}$(A, b)$ 
 \IF {$\psi$ is valid}
\STATE    {\bf return} \(\psi\) 
\change{\ELSE
\STATE  {Halt with failure}}
\ENDIF
\end{algorithmic}
\end{algorithm}

\begin{figure}[t]
\centering
\begin{minipage}{9cm}
\fbox{\includegraphics[width=\textwidth]{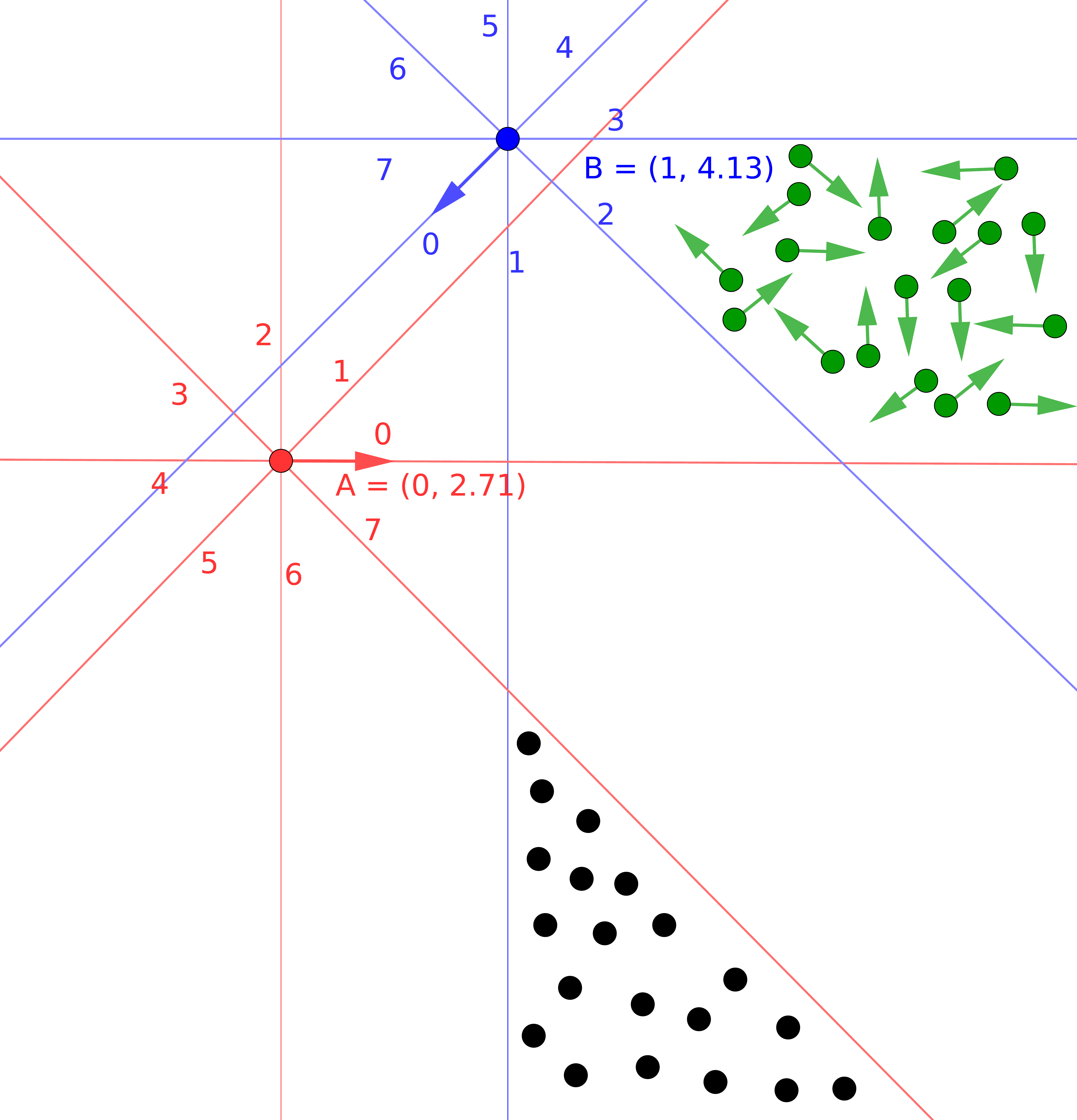}}
\end{minipage}
\caption{Map returned by $StarVars_8$ with generated particles considering a domain with two agents: observer (A) and guided (B), and one goal region; particles representing the goal position are non-oriented dots and those representing the guided agent are the oriented dots.}\label{fig:exemplo}
\end{figure}

The Qualitative Particle Filter (QPF) algorithm is a version of Algorithm \ref{algo:geral} in which the $\texttt{mapping}_{c}$ function is entirely executed by solving the linear programs defined by the set of $StarVars_m$ relations between every pair of objects in the domain (cf. Section \ref{sv}). This procedure is summarised in Algorithms \ref{a2}, where the function $\texttt{buildInequalities}$ uses the data contained in $ Z_c $, in addition to the number of observer agents ($n_o$) and the total number of spatial entities ($n$), to construct a $StarVars_m$ inequality system (linear program) representing the domain. The $\texttt{searchValidModel}$ function returns the $\psi$ world model as a solution of this linear program. Having the possible positions of all the agents and the goal, the $\texttt{releaseParticles}_{c}$ function (Line \ref{release} of Algorithm \ref{algo:geral}) distributes particles in the domain. The output of this process is illustrated in Figure \ref{fig:exemplo}, that shows the domain discretised by the regions formed by the intersections of the $StarVar_8$ sectors of two agents (one observer (A) and one guided (B)) and the particles representing the goal position (single dots) and the guided agent poses (oriented dots).

With the particles distributed, the action selection is accomplished as described in Section \ref{general}. The prediction step in QPF considers rotation and translation actions separately. The former are trivial, as they correspond to rotating $+90^o$, $-90^o$ or $180^o$, plus a Gaussian noise $\delta_t$. Recalling that the guided agent has no motion model, the latter cannot be obtained in a direct way. Instead, we use the qualitative regions generated by $StarVars_m$ as markers for state transitions. More specifically, the algorithm assumes that the action $Move Forward$ is true for each particle until it crosses one of the borders separating adjacent regions, or after a maximum number of steps if moving towards a boundless part of a region. This defines the stopping condition in Line \ref{23} in Algorithm \ref{algo:geral}. As the guided agent $g$ does not have a direct perception of the domain, the update happens based on the perceptions of the observing agents (that inform the coordinator when $g$ arrives at a new region). After the prediction is executed, and the guided agent arrives at a new region, the observer agents resubmit their $StarVars_m$ descriptions obtained through the $\texttt{getDirection}$ function and the update part of the filter starts. 

The $\texttt{update}_{c}$ function of QPF has three steps: the first is a consistency check of the input data, which is performed using the $StarVars_m$ inequality system, considering $Z_c$ as input (Line $ \ref{26}$ of Algorithm \ref{algo:geral}). If this process returns a world model, the input data is consistent and the algorithm continues, otherwise Algorithm \ref{algo:geral} is reinitialised\footnote{The re-initialisation does not always guarantee the returning of a world model, due to sensor noise and the consequent incorrect detection of spatial relations. This is handled by re-executing the procedure until a model is obtained.}. The second step is the calculation of the importance factor $w$ for each particle. Since the entry $Z_c$ contains a new region ($\varrho_{g}$) where the guided agent $g$ is located, it is necessary to get the region $\varrho_p$ for each particle $p$. Let $\Omega_p$ represents the set of $\varrho_p$ locations. With $\Omega_p$ and  $\varrho_p$, it is possible to calculate the probability that each particle is in the same region as the guided agent.  To determine the importance factor $w$ for each particle $p$, the Euclidean distance between $\varrho_g$ and $\varrho_p$ is calculated and $w$ is obtained according to Equation \eqref{eq:w}.
\begin{align} \label{eq:w}
w = \frac{1}{e^{\sqrt{\sum_{i = 1} ^ n (\varrho_ {g} ^ i - \varrho_{p}^i)^2}}}.
\end{align}


Finally, the third (and last) step of the upgrade executes particle resampling, which is done according to $w$. This process continues until the guided agent reaches the goal location.

\subsection{Particle Filter with Qualitative Commands (PFQC) Algorithm}\label{npfgnqi}

Analogously to QPF, the first step of the Particle Filter with Qualitative Commands (PFQC) algorithm is to get the directions (as perceived by every observer agent) by means of the $\texttt{getDirection}$ function. In PFQC, the directions are the angles $\alpha_{ij}^t $ (where $\alpha_{ij}^t \in [0, 2\pi)$) between each observer agent $o_i$ with respect to every domain element $e_j$ at time $t$. Since each agent $o_i$ has knowledge of its own direction $\alpha_i = 0^o$ (that is identical to its orientation $\theta_i$), then $(z_t)_i^j = \alpha_{ij}^t$ at any instant $t$. The second step is the creation of a map, a procedure that is executed by the coordinator agent with the $\alpha_{ij}^t$ data submitted by all observer agents. In order to construct this map, two tasks are performed (as shown in Algorithm \ref{algo:total}): {\em triangulation} and {\em getting the coordinates} of each spatial entity in the domain. Triangulation is a traditional trigonometric technique that, in this work, is responsible for inferring the distances between spatial entities. In turn, the $x_i$ and $y_i$ coordinates of an $ e_i $ entity are obtained by applying the cosine's law (represented by the function \texttt{GetCoordinates} in Algorithm \ref{algo:total}), using the angle between $o_c$ with respect to each  $e_i$, $\alpha_{ci}$, and the distance between the $o_c$ and $e_i$. The world model $\psi$ returned by Algorithm \ref{algo:total} is analogous to that introduced in Section \ref{qpfa}, $\psi = \bigwedge_{i \in \{1, ..., n\}} (x_i, y_i, \theta_{i})$. However, in QPF $\psi$ is obtained as a result of linear programming, while in the present case it is obtained from numeric information.

\begin{algorithm}[b]
\caption{$\texttt{mapping}_{c}$ for PFQC}\label{algo:total}
\begin{algorithmic}[1]
\REQUIRE{\(Z_c, n_o, n \)}
\STATE	$D = $ \texttt{triangulation}$(Z_c, n_o)$ 
\STATE	$X, Y = $ \texttt{GetCoordinates}$(Z_c, D, n)$ 
\STATE	$\psi = <X, Y, \Theta>$ 
\STATE	Return \(\psi\)
\end{algorithmic}
\end{algorithm}

After the mapping task is accomplished, $\texttt{releaseParticles}_{c}$ distributes the particles representing the possible poses of the guided agent (cf. Algorithm \ref{algo:geral}) and the selection of the action to be executed is done by by the $\texttt{chooseAction}_{c}^{g}$ function as explained above (Section \ref{qpfa}). It is only at this point that the $StarVars_m$ agent definition is used in the PFQC algorithm, translating the chosen action into a qualitative relation. After that, the main algorithm follows the prediction-update cycle of the particle filter.

 The prediction of a rotation motion is analogous to that presented in Section \ref{qpfa}. For the translation motion, the predicted location of the particle is defined as a mean value of the distance covered by each particle, assuming a Gaussian distribution. The update part of the particle filter in this case occurs after a maximum time interval ($\tau$), this update rate {\em defines the stopping condition} at Line \ref{23} of Algorithm \ref{algo:geral}. When the stopping condition is reached, the observing agents must resubmit the information about their domain perceptions, obtained through the $\texttt{getDirection}$ function, to the coordinating agent. Then the $\texttt{update}_{c}$ function is executed, as in QPF, in three steps: first, the guided agent executes a moving action, then the importance factor is calculated for each particle and, finally, resampling is executed. The importance factor $w$ is based on a Gaussian function assuming the standard deviation for particle update and the Euclidean distance between $x_i$ and $y_i$ of each particle $p_i$ and the guided agent's new position $x_g$ and $y_g$.

\section{Experiments and Results}\label{results}

In order to evaluate the algorithms proposed in this work, a number of experiments were carried out with virtual humanoid robots, in which four guided navigation methods were tested: two entirely based on $StarVars_m$ (without particle filters) and the two algorithms proposed in this work: QPF and PFQC. 

The first method used $StarVars_m$ with a {\em single instruction}; i.e., the model is the direct application of $StarVars_m$ by the coordinating agent. From the perception received from all observer agents, the coordinator agent infers the direction (angular section), with respect to the guided agent's orientation, in which the goal is located; it then sends a single high-level command to the guided agent. Basically, this method ignores motion, perception, or noise and is used as a baseline experiment. The second $StarVars_m$-only method (called ``$StarVars_m$ with {\em multiple updates}") can be understood as an intermediate definition between the single-command $StarVars_m$ and QPF since, although the notion of particles is not present, a new world model is built with $StarVars_m$ every time the guided agent arrives at a new region.

In addition to the tests with the simulator, two proofs of concept with real humanoid robots were carried out, one for each of the algorithms proposed in this paper (QPF and PFQC), as described in Section \ref{tests_hum}.

\subsection{Tests in the simulator}



Experiments were executed in a Humanoid Robot 2D Simulator proposed in \cite{perico2018humanoid} that allows portability between simulated and real robots.

\begin{figure}[t]
\centering
\begin{minipage}{7cm}
\includegraphics[width=\textwidth]{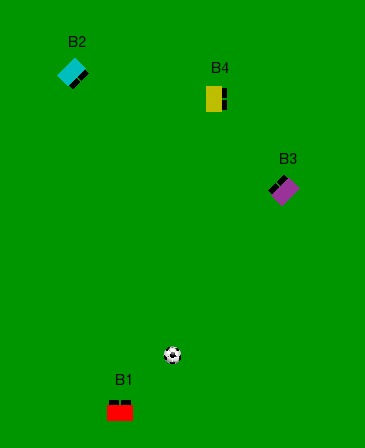}
\end{minipage}
\caption{Snapshot of the simulated domain.}\label{fig:simulation}
\end{figure}

\begin{figure}[t]
\centering
\begin{minipage}{7cm}
\includegraphics[width=\textwidth]{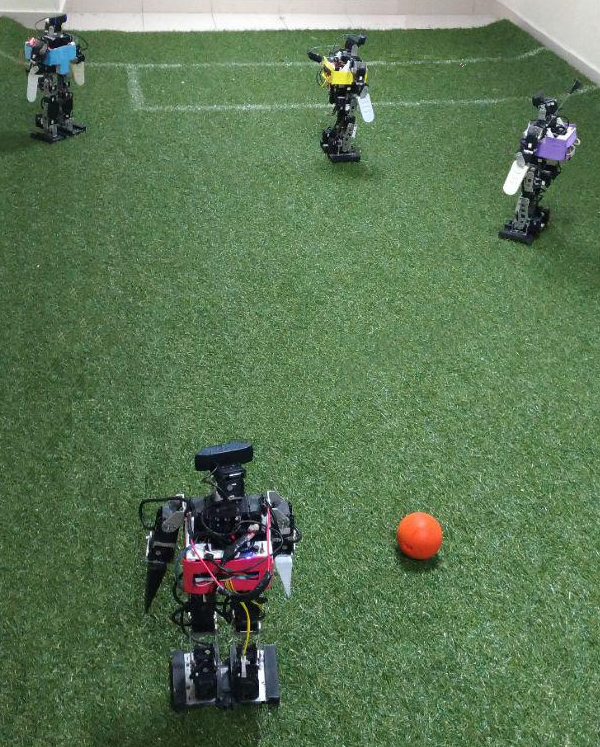}
\end{minipage}
\caption{Snapshot with real robots.}\label{fig:real_robots}
\end{figure}

The main idea of the simulator is to reproduce the processes of vision, control and inertial mesuremente unit (IMU) of $n$ robots. By default, the simulator uses the domain of the Humanoid League of RoboCup Soccer \citep{RoboCup}, which has the scenario of a football field measuring $900 \times 600$ $cm$ (plus $70 cm$ on each side) and humanoid robots as agents. Figure \ref{fig:simulation} shows a scene in the simulator, representing the situation with the real robots depicted in Figure \ref{fig:real_robots}. Errors can be included in the virtual environment in the form of random numbers generated from normal distributions. Among the various types of errors that can be inserted into the virtual robot, the most significant are: error in speed, which causes a faster or slower walk than expected, lateral slip, which causes unwanted lateral movements and rotation error. The simulator also allows uncertainty in the robot's perception, both in the vision processes, and in the IMU data. Gaussian noise with a zero average and standard deviation of $1$ was included in the process, that simulates the computational vision of each virtual robot for perceiving directions. All experiments were carried out with five spatial entities: three observer humanoid robots, a guided humanoid robot and a goal, represented by the ball. The environment in which the robots were embedded had an area of $1000\times 1000 cm$. The translation speed used in the robot during the experiments was $10.0 cm/s$ and the angular velocity was $0.31 rad/s$. All simulated experiments were performed on an Intel$^{\circledR}$ NUC Core i5-4250U 1.30GHz, 8GB SDRAM and 120GB SSD with Linux Operating System, using Ubuntu distribution 16.04. The code was developed in Python, version 2.7. The source code is available at the following URL: \url{https://github.com/danilo-perico/guided_navigation_qsr}.

Each algorithm was tested $100$ times with randomly generated initial positions for all spatial entities, in which a distance of $250 cm$ was respected from each edge of the domain. This setup was submitted to each of the four methods, varying the granularity $m$ for all cases and the value of $\tau$ (measured in seconds) for PFQC. The completion of an episode was defined as the completion of one guided navigation, from an initial position to a goal location. In order to apply the algorithms, the robots were assigned distinct colours, so that simple colour segmentation algorithms could be used in the identification process.

The main factor considered in the analysis was the success rate of executing the task of leading the guided agent to a goal location. Table \ref{success} displays the success rate after 100 episodes, when the guided agent had information about its own orientation (column ``orientation") and when this information was not available (column ``no orientation"). The two basic $StarVars_m$ algorithms were tested with granularity $m=\{8,16\}$, while QPF and PFQC were tested with $m=\{8,16, 32\}$; the maximum time interval $\tau$ of 3 and 6 were used in PFQC.

\begin{table}[b!]
\centering
\caption{Success rate for reaching the goal region}\label{success}
\begin{minipage}{10.5cm}
\begin{tabular}{|l|c|c|c|c|}
\hline
& $m$ & $\tau$ & 
\begin{tabular}[c]{@{}c@{}}orientation 
\end{tabular} & No orientation \\ \hline
\multirow{2}{*}{\begin{tabular}[c]{@{}l@{}}$StarVars_m$\\ (single command)\end{tabular}} 
& $8$  & - & $51\%$ & $32\%$ \\ \cline{2-5} 
& $16 $ &-& $41 \% $ & $20 \% $ \\ \hline
\multirow{2}{*}{\begin{tabular}[c]{@{}l@{}}$StarVars_m$\\ (multiple updates)\end{tabular}}                                               
& $8$  & - & $93\%$ & $28\%$ \\ \cline{2-5} 
& $16 $ &-& $93 \% $ & $9 \% $ \\ \hline
\multirow{2}{*}{QPF$_m$}                                                                
& $8$  & - & $92\%$ & $80\%$    \\ \cline{2-5} 
& $16$ & - & $94\%$ & $97\%$    \\ \cline{2-5} 
& $32 $ &-& $96 \% $ & $93 \% $	 \\ \hline
\multirow{4}{*}{PFQC$_m$}                                                              
& $8$  & $3$ & $91\%$  &  $93\%$ \\ \cline{2-5} 
& $8$  & $6$ & $93\%$  &  $93\%$ \\ \cline{2-5} 
& $16$ & $3$ & $94\%$  &  $92\%$ \\ \cline{2-5} 
& $16$ & $6$ & $92\%$  &  $88\%$\\ \cline{2-5} 
& $32 $ & $6 $ & $94 \% $ & $80 \% $ \\ \hline
\end{tabular}
\end{minipage}
\end{table}

Table \ref{success} shows that, with the exception of $StarVars_m$ with single command, the other three methods were successful in the task of guiding a robot (with known orientation, but no perception) in at least $91\%$ of the cases, regardless of the $m$ or $\tau$ values. In the case where the guided agent had no information about its orientation, the purely qualitative methods tend to fail, since $StarVars_m$ returns the first valid world model obtained, which (without the knowledge of the robot's orientation) will always pick orientation $0^o$. In contrast, both QPF and PFQC present high success rates in this case, as a result of the probabilistic filtering procedure applied.

Tables \ref{sem_orient} and \ref{com_orient} present the statistics (considering {\em only} the successful cases) of the tests executed. These tables show the mean values (averaged over 100 episodes, with standard deviation in parenthesis) of the number of commands issued to reach the goal region ($\#$ instr.), the processing time taken to infer each command (proc. time) and the path size. These values were grouped into those related to the guided agent {\em without} information about its orientation (Table \ref{sem_orient}) and those obtained when the agent had information about its own orientation (Table \ref{com_orient}).

\begin{table}[t]
\centering
\caption{Successful experiments {\bf without} orientation (average of 100 episodes, with standard deviation in parenthesis)}
\label{sem_orient}
\begin{tabular}{|l|c|c|c|c|c|}
\hline
& $m$  & $\tau$ & \begin{tabular}[c]{@{}c@{}}\# instr. \\ \end{tabular} & \begin{tabular}[c]{@{}c@{}}proc. time\end{tabular} & \begin{tabular}[c]{@{}c@{}}path \\ size\end{tabular}   \\ \hline
\multirow{2}{*}{\begin{tabular}[c]{@{}l@{}}$StarVars_m$\\ (single instr.)\end{tabular}} 
& $8$  & - & $1$    & $6.61$ $(9.68)$   & $80.11$ $(45.51)$  \\ \cline{2-6} 
& $16$ & - & $1$    & $21.65$ $(19.92)$ & $70.09$ $(38.03)$   \\ \hline
\multirow{2}{*}{\begin{tabular}[c]{@{}l@{}}$StarVars_m$\\ (multiple updates)\end{tabular}}                                                               
& $8$  & - & $6.92$ $(9.79)$ & $4.04$ $(3.95)$   & $115.35$ $(149.77)$  \\ \cline{2-6} 
& $16$ & - & $5.56$ $(4.16)$ & $27.32$ $(23.83)$ & $132.28$ $(163.19)$  \\ \hline
\multirow{2}{*}{QPF$_m$}                                                                
& $8$  & - & $7.53$ $(6.33)$  & $6.40$ $(4.00)$ & $240.75$ $(300.50)$    \\ \cline{2-6} 
& $16$ & - & $10.17$ $(5.58)$ & $6.04$ $(3.84)$ & $172.39$ $(141.12)$  \\ \cline{2-6} 
& $32$ & - & $18.92$ $(9.11)$ & $6.11$ $(3.69)$ & $171.31$ $(116.22)$    \\ \hline
\multirow{4}{*}{PFQC$_m$}                                                              
& $8$  & $3$ & $12.58$ $(8.07)$ & $3.07$ $(0.04)$ & $120.32$ $(68.01)$   \\ \cline{2-6} 
& $8$  & $6$ & $9.99$ $(8.34)$  & $3.07$ $(0.06)$ & $114.27$ $(71.64)$    \\ \cline{2-6}
& $16$ & $3$ & $16.40$ $(8.91)$ & $3.07$ $(0.02)$ & $149.12$ $(67.15)$   \\ \cline{2-6} 
& $16$ & $6$ & $11.69$ $(5.99)$ & $3.07$ $(0.05)$ & $135.53$ $(44.88)$    \\ \cline{2-6}
& $32$ & $6$ & $17.10$ $(9.64)$ & $3.07$ $(0.06)$ & $176.16$ $(130.04)$   \\ \hline
\end{tabular}
\end{table}

\begin{table}[t]
\centering
\caption{Successful experiments {\bf with} orientation (average of 100 episodes, with standard deviation in parenthesis)}
\label{com_orient}
\begin{tabular}{|l|c|c|c|c|c|}
\hline
& $m$  & $\tau$ & \begin{tabular}[c]{@{}c@{}}\# instr. \\ \end{tabular} & \begin{tabular}[c]{@{}c@{}}proc. time\end{tabular} & \begin{tabular}[c]{@{}c@{}}path \\ size\end{tabular}   \\ \hline
\multirow{2}{*}{\begin{tabular}[c]{@{}l@{}}$StarVars_m$\\ (single instr.)\end{tabular}} 
& $8$  & - & $1$ & $3.52$ $(2.04)$    & $69.38$ $(42.78)$    \\ \cline{2-6} 
& $16$ & - & $1$ & $49.60$  $(76.70)$ & $80.13$ $(26.20)$   \\ \hline
\multirow{2}{*}{\begin{tabular}[c]{@{}l@{}}$StarVars_m$\\ (multiple updates)\end{tabular}}                                                               
& $8$  & - & $7.86$ $(7.29)$  & $3.21$ $(1.42)$   & $122.98$ $(135.66)$ \\ \cline{2-6} 
& $16$ & - & $13.14$ $(8.93)$ & $35.80$ $(43.18)$ & $150.79$ $(89.27)$  \\ \hline
\multirow{2}{*}{QPF$_m$}                                                                
& $8$  & - & $5.44$ $(4.72)$ & $6.55$ $(4.39)$ & $124.11$ $(189.35)$    \\ \cline{2-6} 
& $16$ & - & $8.78$ $(6.01)$ & $6.09$ $(4.14)$ & $113.18$ $(66.50)$    \\ \cline{2-6} 
& $32$ & - & $17.94$ $(10.81)$& $6.03$ $(3.68)$ & $179.80$ $(224.15)$   \\ \hline
\multirow{4}{*}{PFQC$_m$}                                                              
& $8$  & $3$ & $16.20$ $(10.70)$ & $3.06$ $(0.05)$ & $93.03$ $(59.95)$ \\ \cline{2-6} 
& $8$  & $6$ & $8.24$ $(5.74)$   & $3.06$ $(0.08)$ & $102.40$ $(56.78)$  \\ \cline{2-6} 
& $16$ & $3$ & $15.67$ $(10.64)$ & $3.07$ $(0.05)$ & $124.82$ $(65.37)$  \\ \cline{2-6} 
& $16$ & $6$ & $10.68$ $(6.54)$  & $3.06$ $(0.06)$ & $122.25$ $(68.83)$   \\ \cline{2-6}
& $32$ & $6$ & $12.98$ $(9.11)$  & $3.06$ $(0.06)$ & $140.63$ $(87.10)$ \\ \hline
\end{tabular}
\end{table}

Tables \ref{sem_orient} and \ref{com_orient} show that, among the successful tests, the shortest paths were obtained by the single instruction $StarVars_m$. However, it is worth pointing out that this was the algorithm that was successful in less than $50\%$ of the cases (cf. Table \ref{success}) and that these cases happened when the goal was exactly in the direction set by the first command issued by the algorithm. Therefore, in these cases, the robot only had to move in a straight line. In most successful cases represented in Tables \ref{sem_orient} and \ref{com_orient} the agents described a trajectory that was greater than the optimal path (due to the qualitative nature of the commands).

$StarVars_m$ with multiple updates achieved a success rate of $ 93 \% $ for both granularities ($m=8$ and $m=16$), when the information about the robot's orientation was available (Table \ref{success}). However, when checking the path taken in this case (Table \ref{com_orient}) it can be noted that, for $m = 16$, the path size is similar to the path obtained by QPF$_{16}$, but the latter presented a much faster processing time per instruction. This suggests that, when the guided robot has information about its own orientation, $StarVars_8$ with multiple updates could be a suitable choice of algorithm. However, for $ m = 16 $, the processing time of $StarVars$ is prohibitively high, in contrast to the other the three methods that obtained a success rates above $91\%$.

In general, the average path sizes are shorter for smaller $m$ (i.e. coarser granularity), since these cases produce larger areas and, therefore, larger (and, thus, more accessible) goal regions. The results also show that the number of instructions is directly proportional to the value of $ m $ for both, $StarVars_m $ with multiple updates and QPF algorithms. In the case of PFQC, the number of instructions is inversely proportional to the refresh rate $\tau$. It can be noted that, for QPF and PFQC, the average processing time is constant and independent from $ m $ or $\tau$. For both $StarVars_m $ methods without particle filtering, the inference time tends to increase with respect to $ m $. 

Analysis of Variance (ANOVA) with Welch test \citep{welch,mcdonald_welch_anova} was performed on the path sizes (Tables \ref{com_orient} and \ref{sem_orient}), assuming a significance level of 0.05. For cases with a known orientation information, the three models with a success rate greater than $90\%$  were considered in this analysis; they were: $StarVars_{m}$ (multiple updates), QPF$_{m} $ and PFQC$_{m}^{6}$ (for $m = 8 $ and $m = 16 $). For $ m = 8 $ the results of this analysis suggest that the size difference of the paths generated was not significantly distinct. On the other hand, for $ m = 16 $, the variance analysis shows that at least one model generated a significantly different path length with respect to the others.  Games-Howell post-hoc test \citep{games-howell} was used to identify that $StarVars_{16}$ (multiple updates) generated significantly larger paths than the other methods tested. There was no significant difference between QPF$_{16}$ and PFQC$_{16}$. The same analysis was conducted for the cases of a guided agent without orientation information for the two models that had a success rate greater than $80\%$ in Table \ref{success}: QPF$_{m}$ and PFQC$_{m}$ (with $\tau = 6)$. The results allow the conclusion that QPF$_{m}$ (for both granularities tested) is worse than PFQC$_{16}$ (with $\tau = 6)$.

\begin{figure}[t!]
\centering
\subfigure[Get direction.]
{\resizebox*{7cm}{!}{\includegraphics{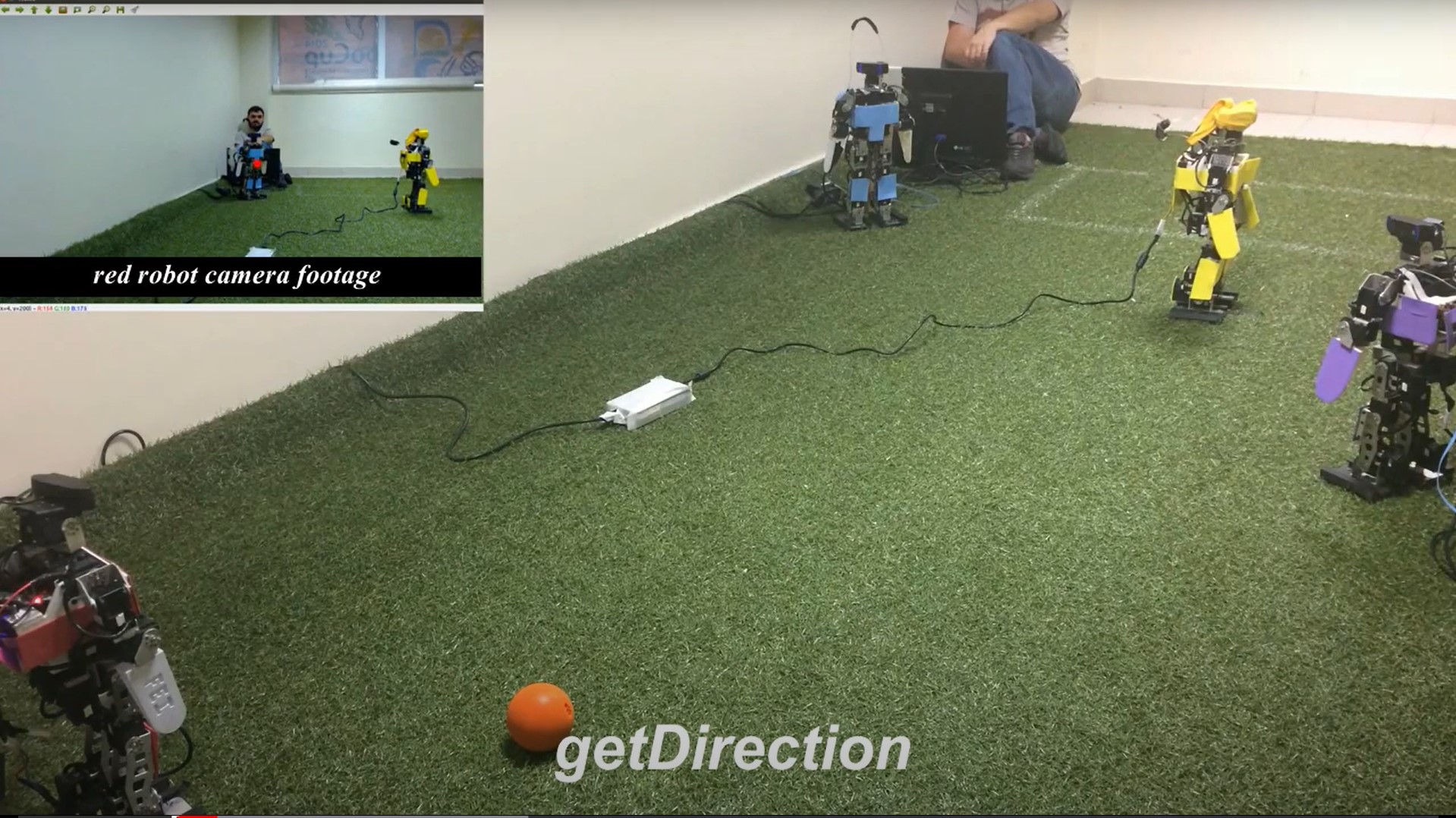}}}
\subfigure[Release particles.]
{\resizebox*{7.2cm}{!}{\includegraphics{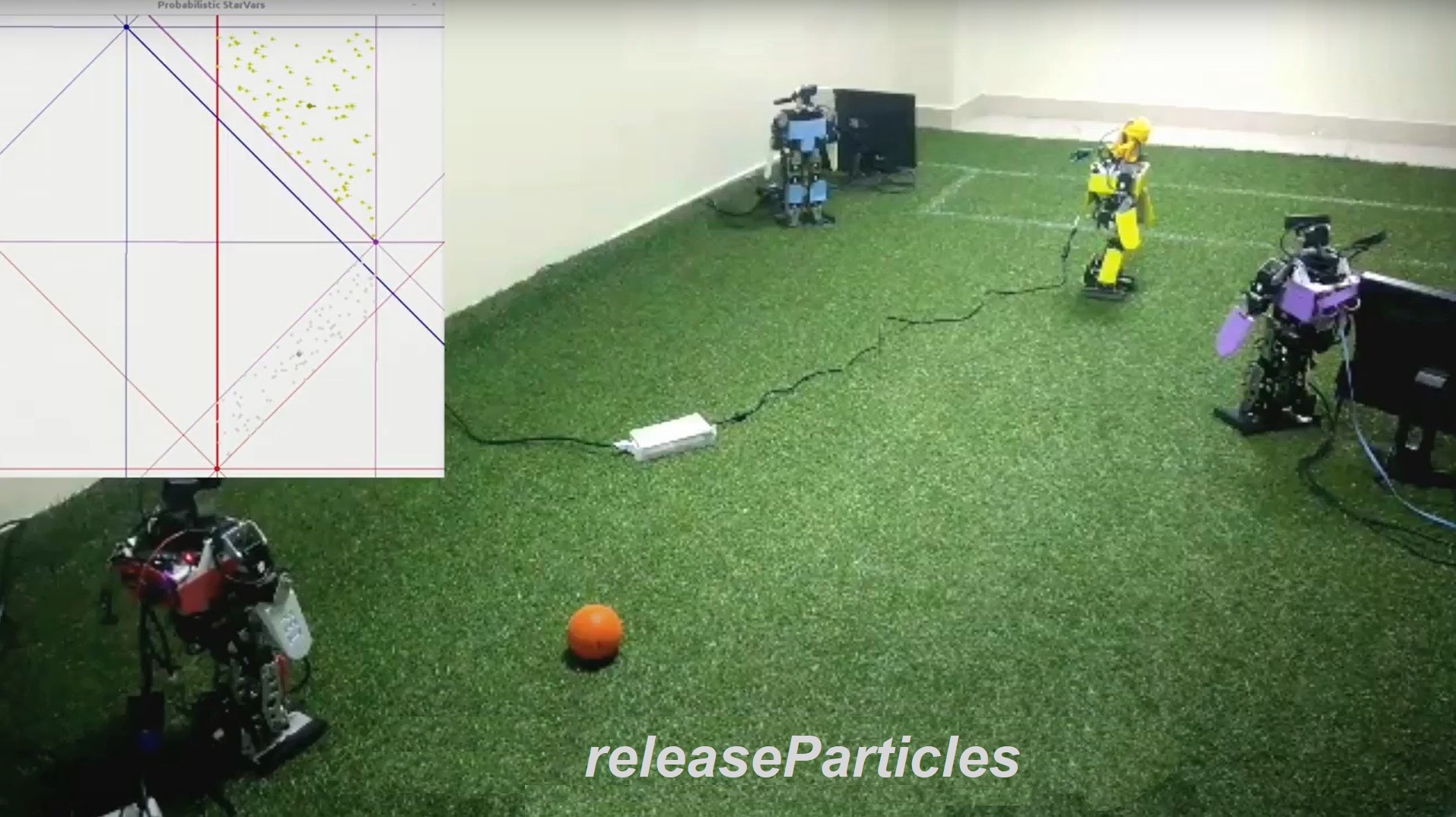}}}
\subfigure[Choose action.]
{\resizebox*{7cm}{!}{\includegraphics{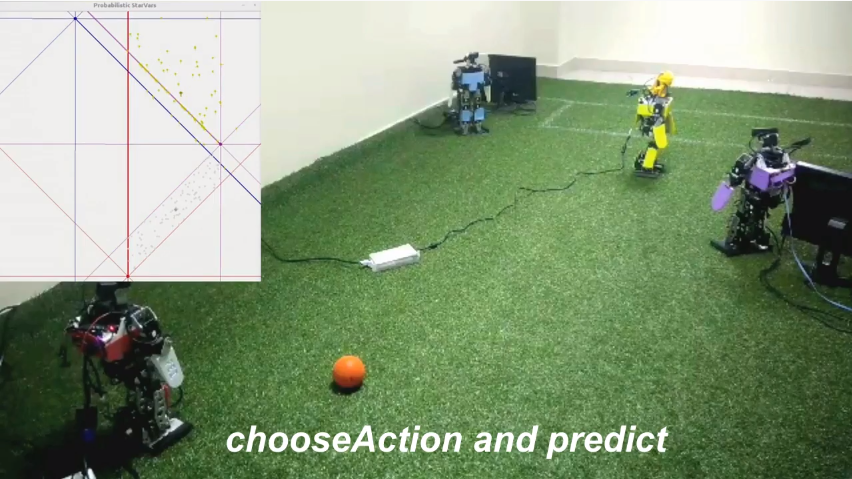}}}
\subfigure[New observations.]
{\resizebox*{7cm}{!}{\includegraphics{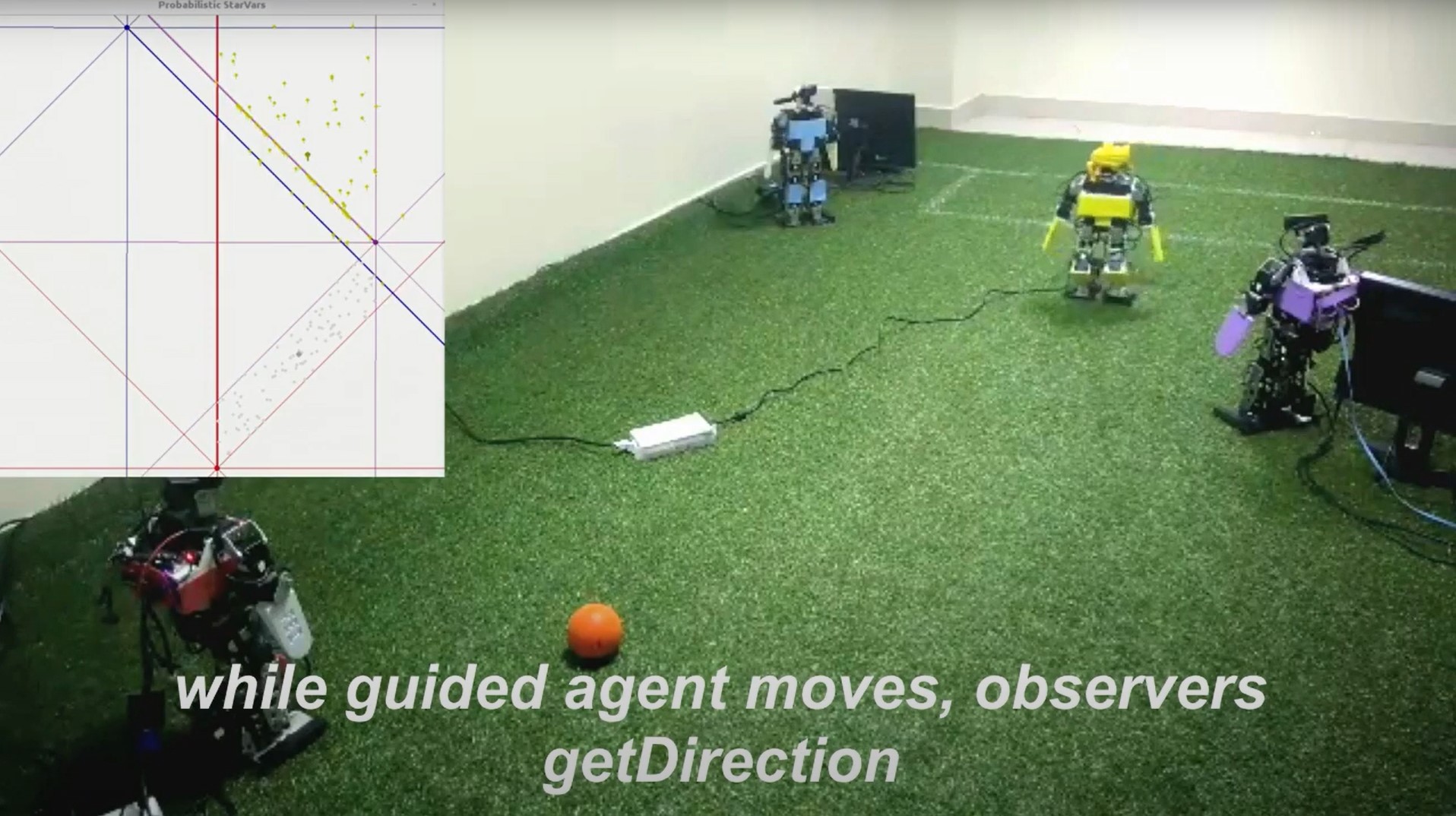}}}
\subfigure[Update particles.]
{\resizebox*{7cm}{!}{\includegraphics{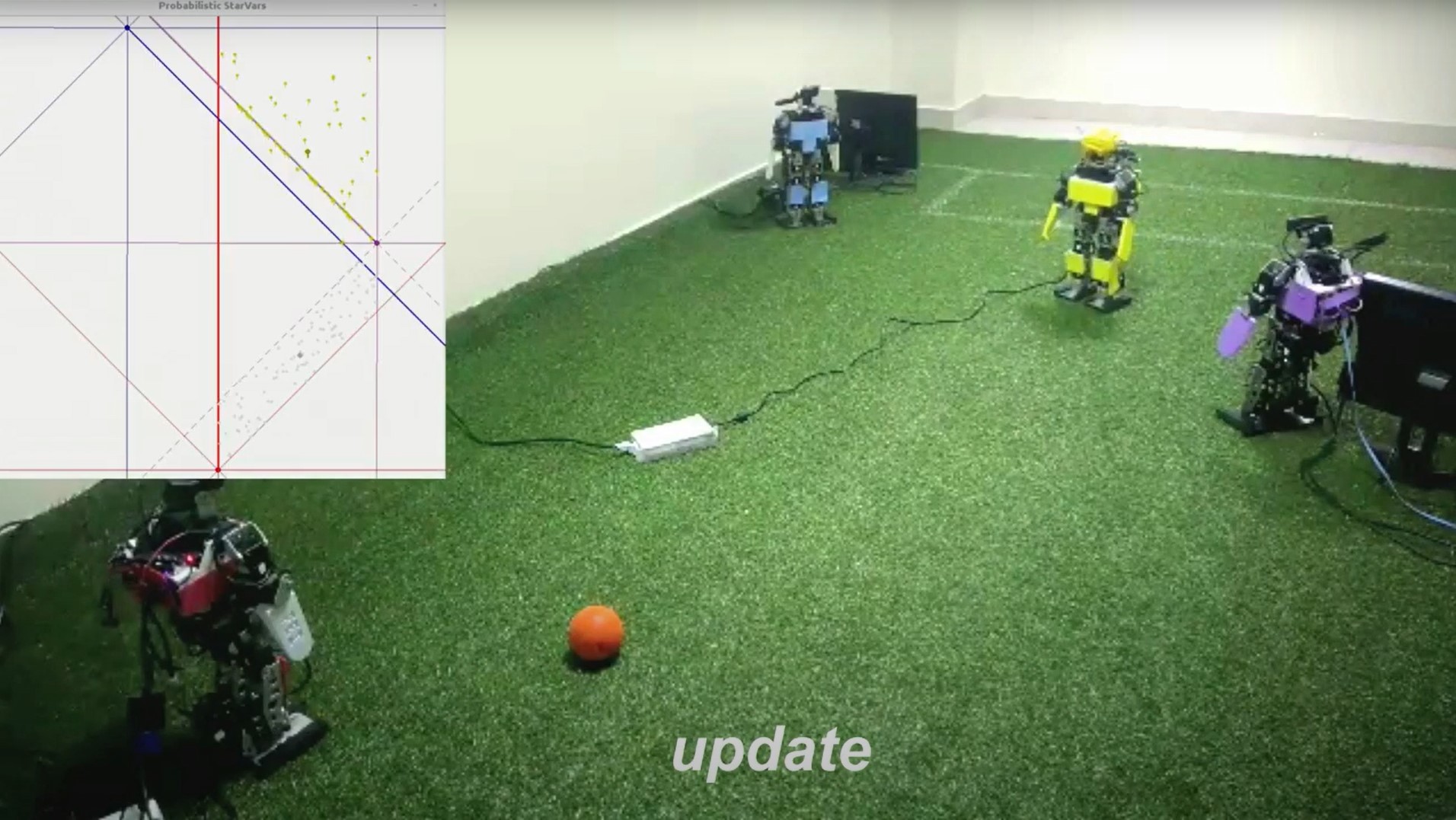}}}
\subfigure[New loop iteration.]
{\resizebox*{7cm}{!}{\includegraphics{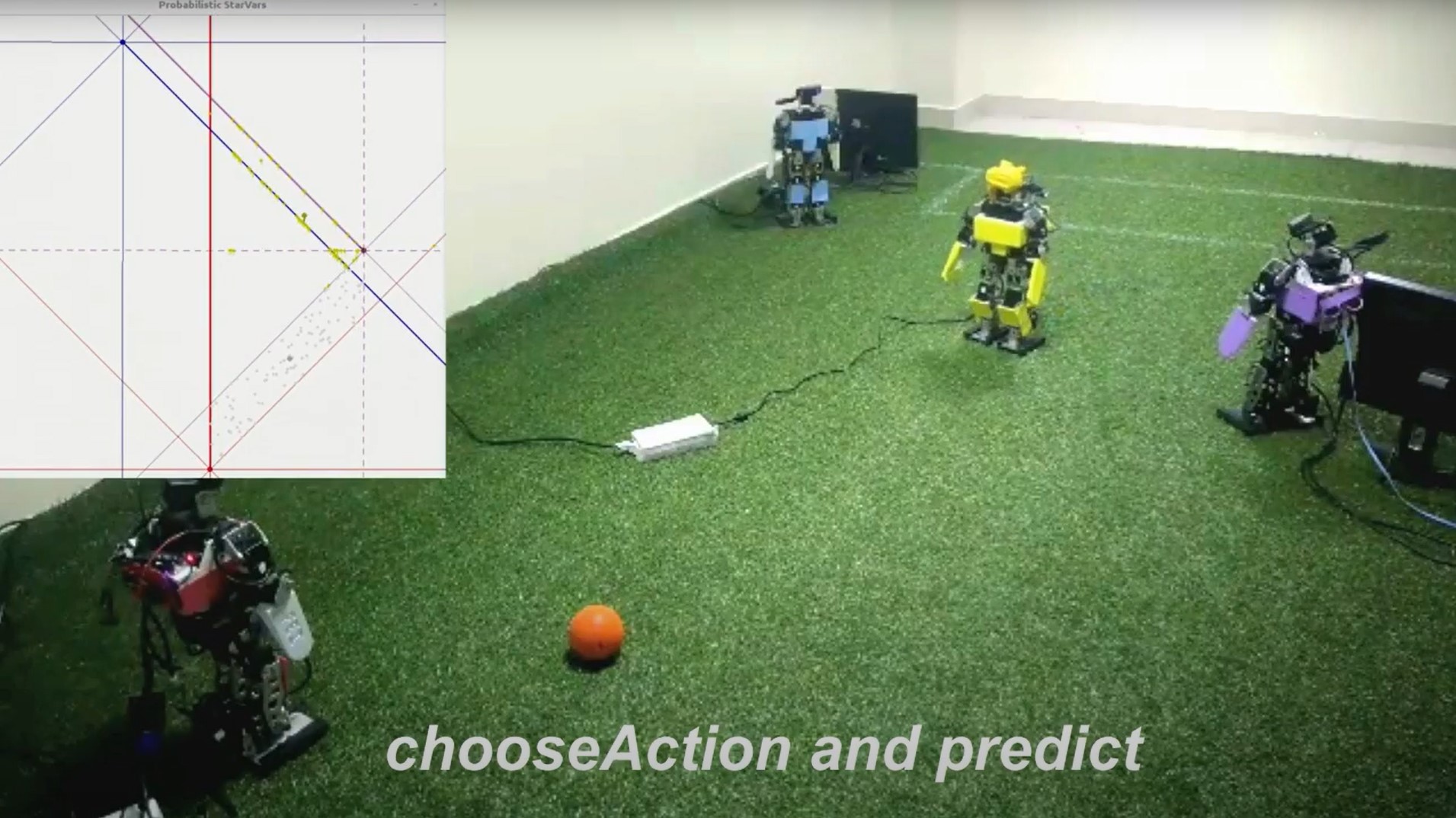}}}
\caption{Example of guiding a ``blind-folded" robot using QPF$_8$ (best seen in colour).}
\label{fig:real_qpf8}
\end{figure}

\begin{figure}[t!]
\centering
\subfigure[Get direction.]
{\resizebox*{7cm}{!}{\includegraphics{state1.jpg}}}
\subfigure[Release particles.]
{\resizebox*{7cm}{!}{\includegraphics{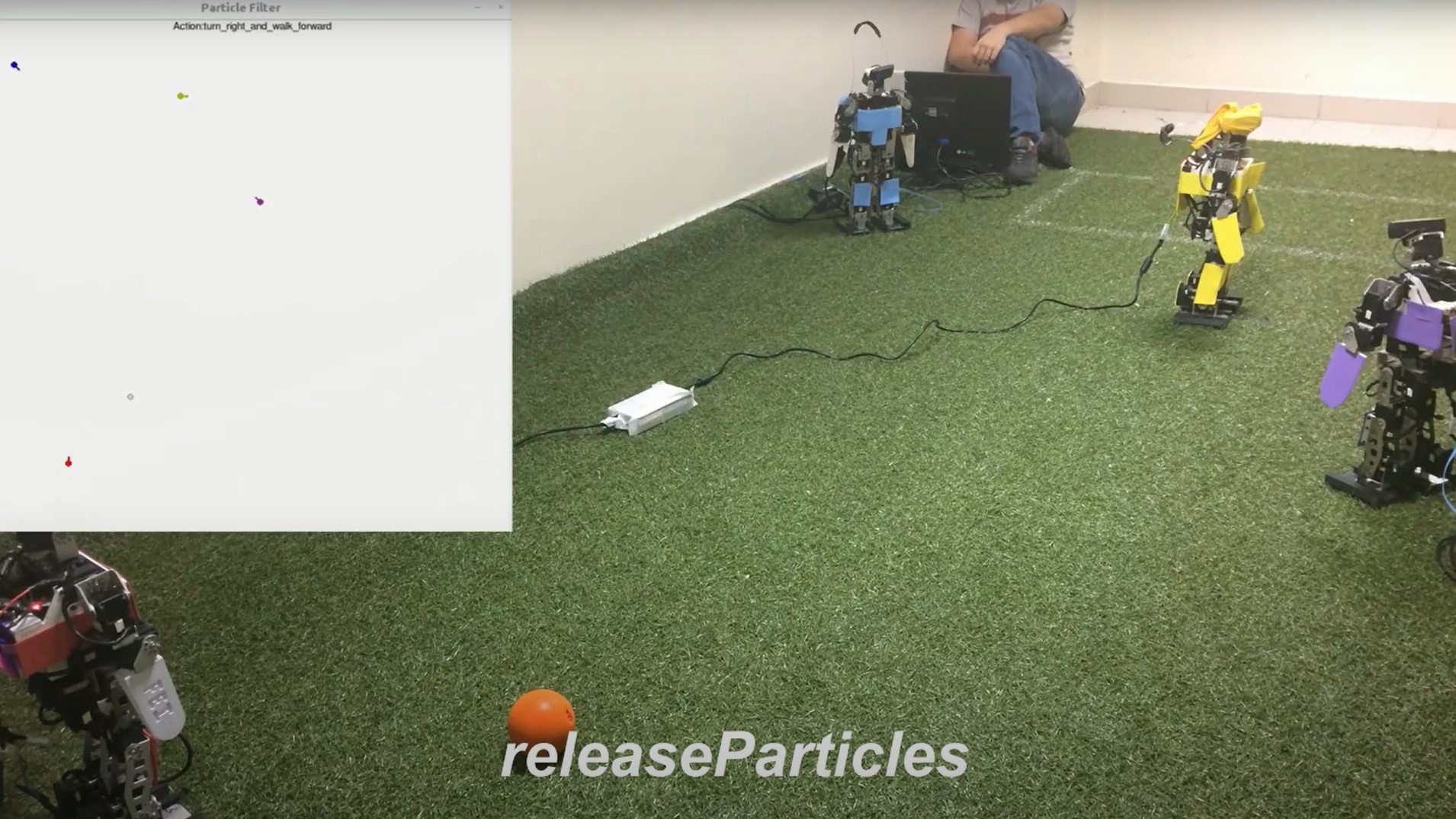}}}
\subfigure[Choose action.]
{\resizebox*{7cm}{!}{\includegraphics{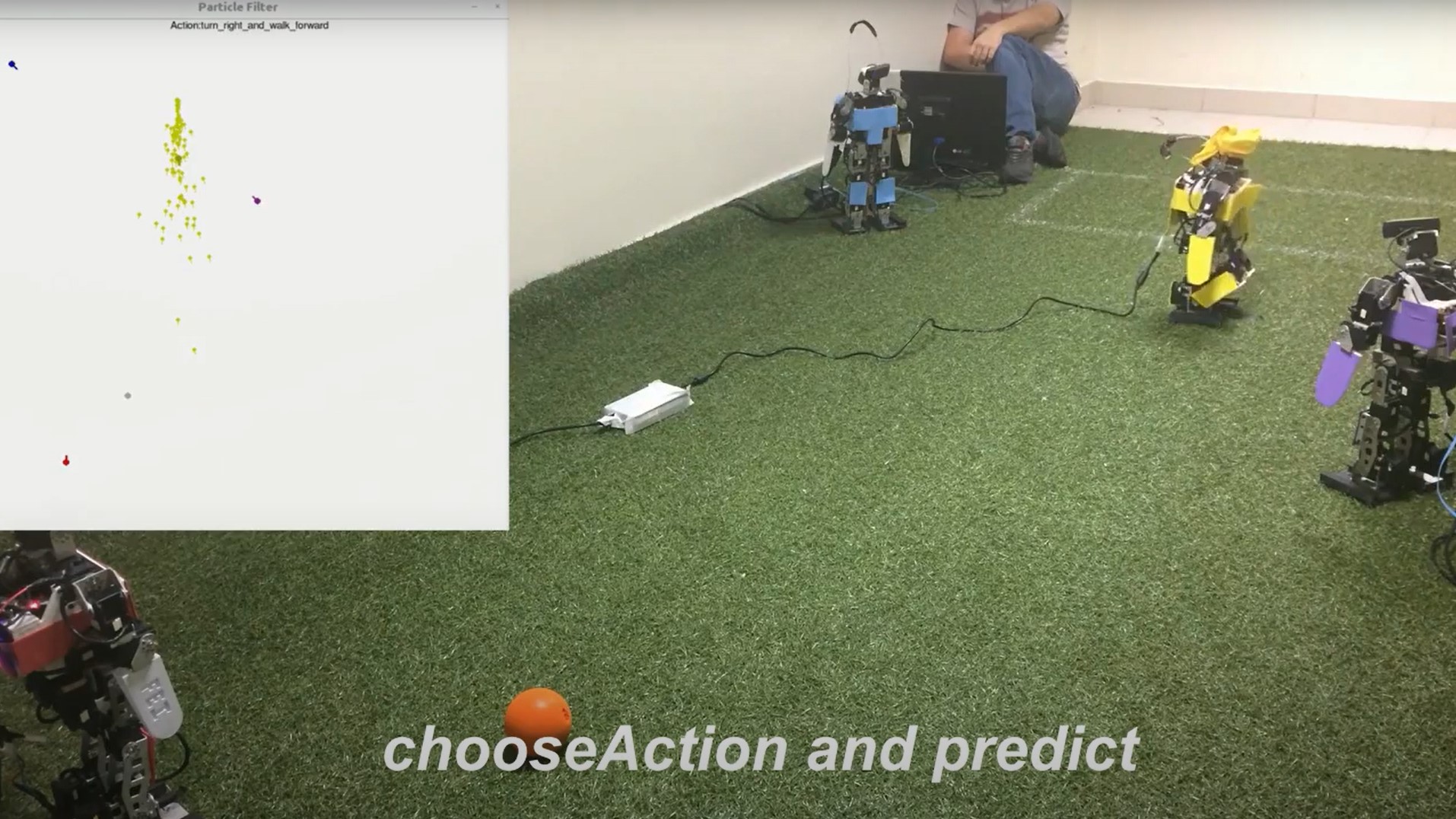}}}
\subfigure[New observations.]
{\resizebox*{7cm}{!}{\includegraphics{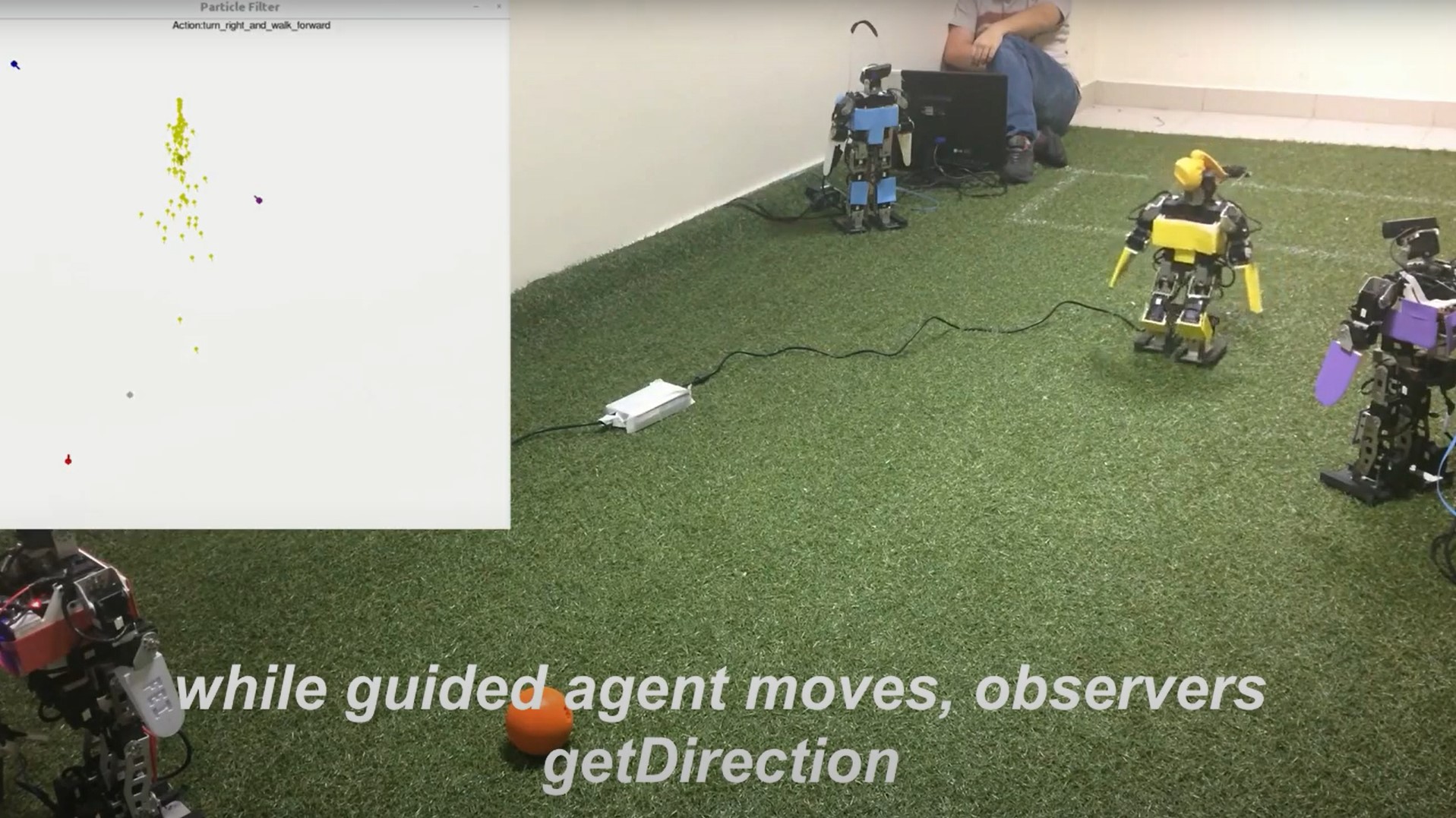}}}
\subfigure[Update particles.]
{\resizebox*{7cm}{!}{\includegraphics{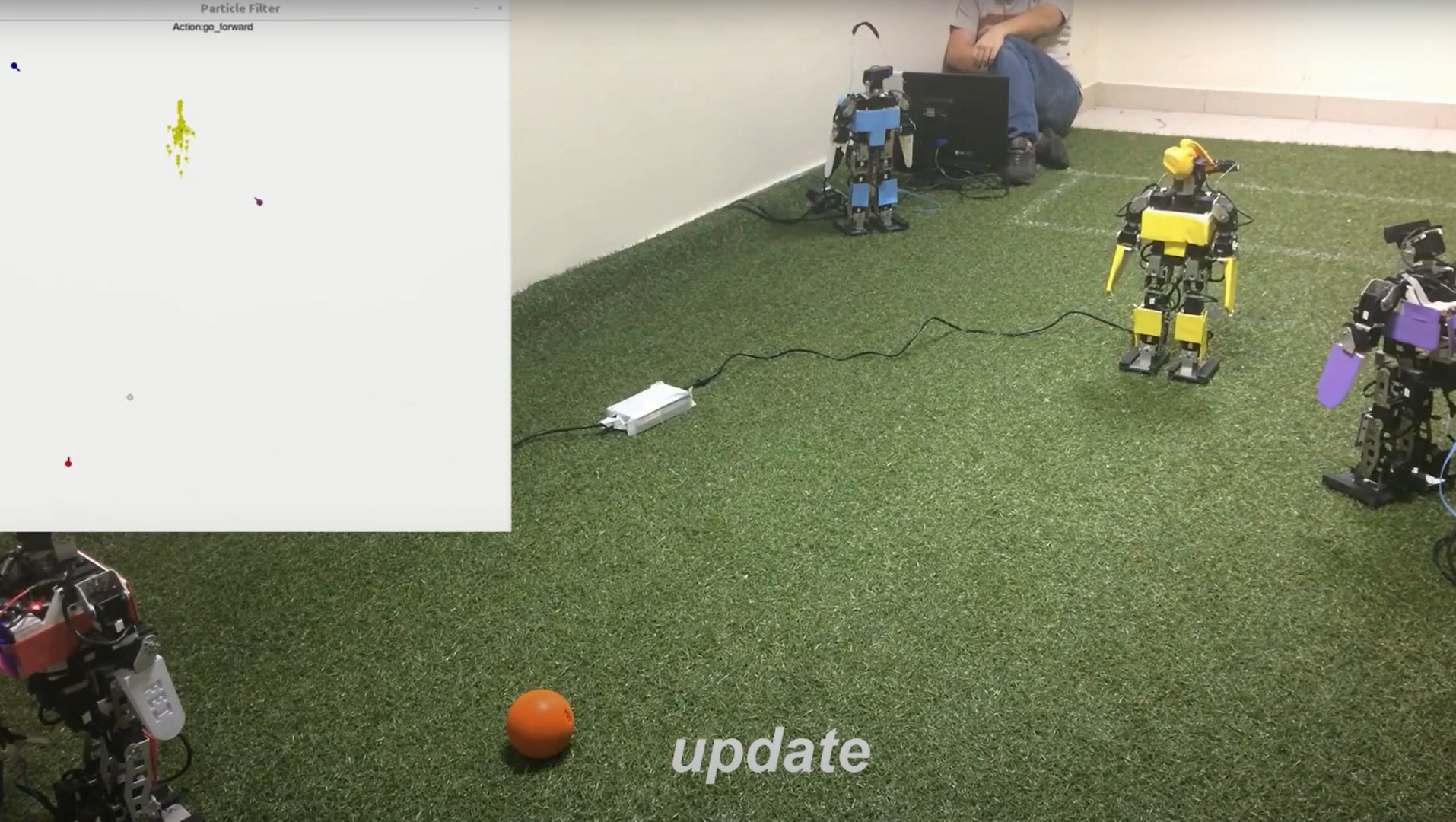}}}
\subfigure[New loop iteration.]
{\resizebox*{7cm}{!}{\includegraphics{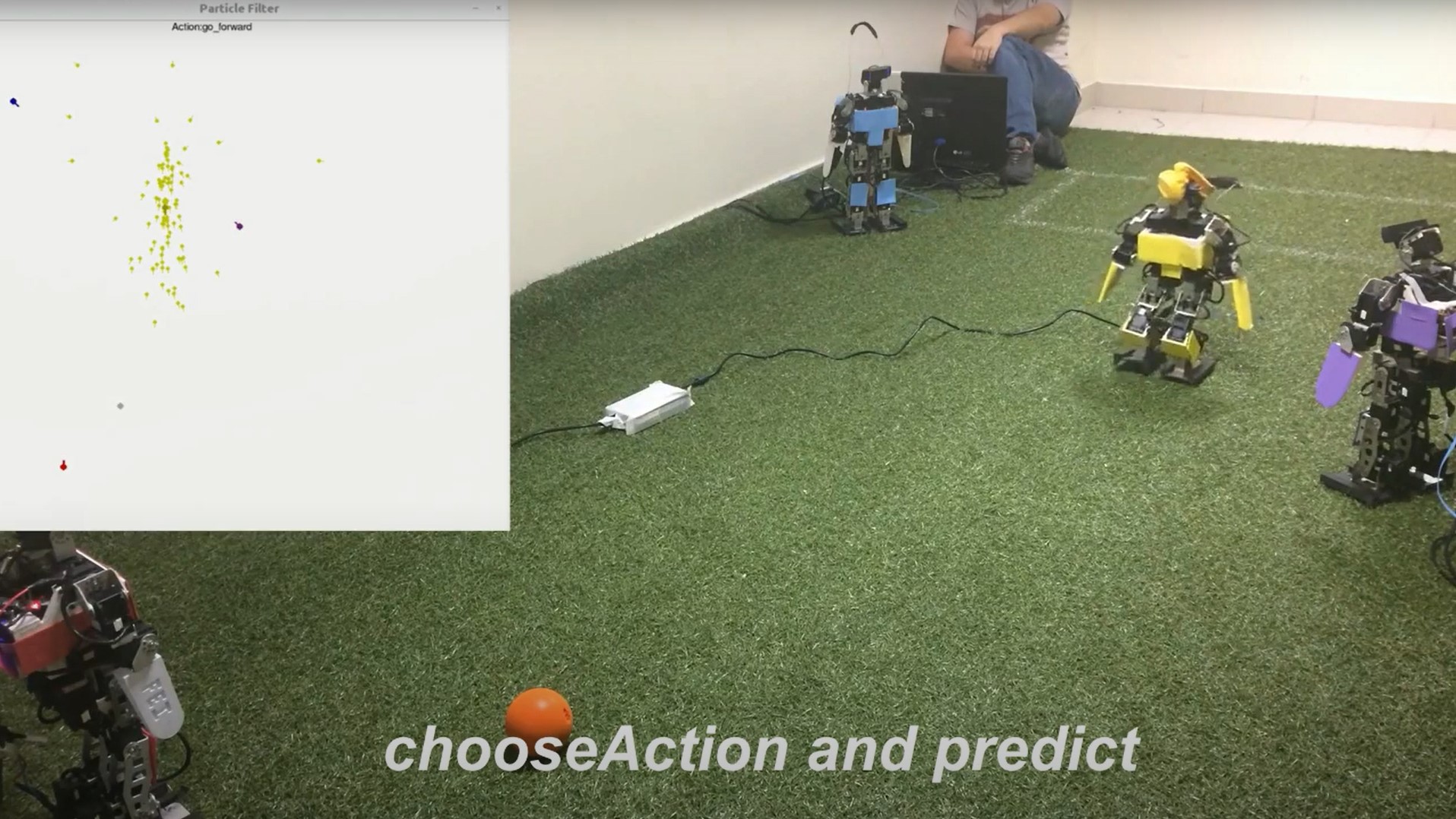}}}
\caption{Example of guiding a ``blind-folded" robot using PFQC$_{16}$ (best seen in colour).}
\label{fig:real_pfqc16}
\end{figure}

\subsection{Tests with real humanoid robots}\label{tests_hum}

A proof of concept was performed applying the methods QPF and PFQC on a team of real humanoid robots (Figures \ref{fig:real_qpf8} and \ref{fig:real_pfqc16}). These tests were conducted with three observer robots (red, blue and magenta) and one guided agent (yellow, ``blind folded" robot), assuming a fixed $m = 8$ for QPF and $m = 16$ for the PFQC (with $\tau = 6$). The guided robot had always information about its own orientation obtained from its IMU. 

The robots and the ball were uniquely identified by their colours. Object recognition was performed by colour segmentation using the OpenCV\footnote{https://opencv.org/} library, in which the $\texttt{moments}$ function was used to identify the geometric centre of the segmented image. The vision of the humanoid robots was implemented to always align the camera to the geometric centre of the object of interest (which was defined as the guided robot). Robots obtained the head steering information by reading the position of the servomotor that controls the pan movement of the camera. 

Each QPF and PFQC were run $3$ times in this domain, successfully leading the guided agent to the goal. \change{Figures \ref{fig:real_qpf8} and \ref{fig:real_pfqc16} show the main steps of the algorithms investigated in this paper applied on real robots. It is worth observing how the distribution of particles in QPF (as well as the related processes of action choice and particle update) is bound by the regions defined by $StarVars_m$, as shown in Figures 6(b)--6(e); whereas in PFQC, the particle distribution has no boundaries (Figures 7(b)--7(e))}. A full experimental evaluation of the algorithms proposed in this work in a real robot environment, however, was left as future research.

\section{Related Work}\label{rw}

The research presented in this paper is inline with the work on Collaborative SLAM (CSLAM) \citep{mourikis}. However, in this work we concentrated on analysing the use of qualitative information in a guided navigation scenario of a sensory deprived robot, an issue that has not received much attention in the past. In general CSLAM runs exclusively with numerical data \citep{Cadena16tro-SLAMfuture}, in contrast to the qualitative information used in the present paper. A similar comment can be can be made with respect to the work reported by \cite{7017662}, that describes a collaboration between a quadcopter guiding a ground robot to achieve a goal location in a search and rescue scenario. The motivation for such development lies in the fact that robots have been constantly used in various disaster situations, but remotely controlled by trained professionals. One of the problems with using manual control of robots is that the operator usually has to send very low level commands to the robot, which makes the rescue slow and creates the need of various operators for each robot. We believe that the methods using qualitative representations investigated in this paper contribute positively to the solution of this issue.

The literature about controlling a robot with high-level routines (or commands), inspired by natural language (or qualitative) terms, is very extensive in time and scope  \citep{LEVITT1990305,Gribble1998,Tellex:2006,Cangelosi2016,8778650,8675641,Nikolaidis:2018}, as recently overviewed in \citep{Mavridis:2015,8298518,liu19}. However, most of the research on this subject were not constructed upon modern qualitative spatial calculi, which provides rigorous mathematical foundations for the development of spatial representation and reasoning tools.

More akin to the qualitative aspect of the work presented in this paper, \cite{OPRAMoratz06} demonstrates how to integrate local knowledge within a qualitative direction calculus called OPRA \citep{OPRA12,opra-composition} using the four-legged robot AIBO. In the tests reported in that paper, the robot was able to distinguish between colours of simple objects and decide their relative location with respect to its egocentric reference frame. OPRA was applied to navigation tasks through street networks in \citep{LueckeEtAL2011}, where the cognitive adequacy of this representation was also investigated. A similar idea was followed in \citep{Kreutzmann_Wolter_Dylla_Lee_2013,cosy:R3-R4-sailaway-aisb} to formalise the Maritime Navigation Collision Prevention Rules. According to \cite{cosy:R3-R4-sailaway-aisb}, compliance with official shipping rules is essential to ensure safety in the maritime domain. Thus, qualitative decision support systems are of interest because they are capable to represent a portion of the human conceptualisation of these rules. However, in contrast to the work presented in this paper, the authors did not develop a way of guiding an autonomous agent using the qualitative information embedded in the rules. OPRA was also used in \cite{HomemPSBM17} and \cite{Bianchi:2018} in a case-based reasoning system for retrieving and reusing qualitative cases for situations involving multiple robots. That work, however, did not assume any probabilistic filtering method and, thus, is prone to sensor and motor noise.

 There are several options of qualitative representations that allow egocentric information to be expressed, such as the Double-Cross Calculus \citep{Freksa92}, the Ternary Point Configuration Calculus (TPCC) \citep{Moratz03} and the Interval Occlusion Calculus (IOC) \citep{santos15,martin17}, among others \citep{chen_cohn_liu_wang_ouyang_yu_2015}. Although $StarVars$ is a generalisation of the directional information contained in these formalisms, there are additional elements in these calculi (such as the representation of occlusion and viewpoint change in IOC) that could lead to interesting avenues for future research, if combined with the algorithms proposed in this paper.

A method was proposed in \citep{Santos09,DBLP:conf/bracis/PereiraCSM13,valquiria16} to solve the robot self-location problem by combining qualitative representation with Markov localisation procedure. This method was implemented as a qualitative-probabilistic approach in which a robot can localise itself within qualitatively distinct regions of space by observing objects, their shadows and occlusions. A Bayesian filter handles sensor uncertainty much in the same way the particle filter applied in this work does. But the similarity ends here, as the probabilistic filter was not used in \citep{Santos09,DBLP:conf/bracis/PereiraCSM13,valquiria16} to generate actions to the robot, as reported in the present paper. However, none of these developments included the automatic generation of maps from unknown environments using qualitative spatial reasoning methods. This issue was tackled in \citep{wall10,MCCLELLAND201673}. In \citep{wall10} a topological map of the environment is constructed by representing hallway junctions in terms of their relative cardinal orientations. \cite{MCCLELLAND201673} investigate a qualitative relational mapping and navigation for planetary rovers that uses the double-cross calculus \citep{Freksa92} to represent a qualitative map and Monte-Carlo simulations to generate measurements of detected landmarks. Although the work reported in \citep{MCCLELLAND201673} assumes distinct methods from those used in this work, that development is based on ideas that are grounded on the same fundamental concepts to those underlying the present paper.

In a broader sense, this work is related to the symbol grounding problem \citep{HARNAD1990335}, and its robotics counterpart sometimes called {\em Anchoring} \citep{CoradeschiSaffiotti.ras03}, that is the problem of connecting symbolic representations to sensor data. We believe that the methods proposed in this paper could contribute with possible solutions to the issue of grounding situation models to conversational agents \citep{Tellex_2011,Mavridis07} and to grounding sensor data to relations in the qualitative modelling of 3D trajectories \citep{vanDerWeghe06,Iliopoulos14,Mavridis:2015}.

\section{Conclusion}

This paper described solutions to guided navigation of a sensory deprived agent, with no fully specified motion model, using observations from various agents in the domain. Two qualitative-probabilistic algorithms were proposed to this end, assuming commands generated based on qualitative relations. The first algorithm, called Qualitative Particle Filter - QPF (Section \ref{qpfa}), defines a particle filter developed upon the qualitative spatial calculus for relative directions called $StarVars$. In this case, motion prediction and particle weight updating are performed based on the qualitative regions generated by combining the distinct sectors of a discretisation of the space around each observer. The second algorithm, called Particle Filter with Qualitative Commands - PFQC (Section \ref{npfgnqi}), runs a particle filter with numerical direction and orientation data to infer commands that are discretised into qualitative relations to guide the sensory deprived robot. These algorithms were tested in a simulated environment and the results obtained were also compared with two purely qualitative navigation methods entirely based on $StarVars$ (one assuming the generation of a single command, and the other executing multiple updates of $StarVars$ world model). In addition, QPF and PFQC were tested with real humanoid robots as a proof of concept.

The results obtained suggest that $StarVars$ with multiple updates was capable of providing appropriate commands for guiding the agent to the goal, when the guided agent had information about its orientation. If such information is not present, only QPF and PFQC were capable of guiding a sensory deprived robot to a goal location efficiently. However, PFQC with granularity m = 16 provided the best trade off between quality of the solution and processing time.

In this work, the coarsest representation considered is $StarVars_8$. In particular in context of the particle filter variants and their high success rates reported in this paper, it would be interesting to see how much coarser a representation could be, while still guiding the sensory deprived robot to the goal location. This analysis is a task for future research.

As a future work we also intend to consider moving objects interacting with the guided agent in the problem. This interaction could be formalised in terms of the relative trajectories executed by the moving robots using the Qualitative Trajectory Calculus (QTC) \citep{vanDerWeghe04,vanDerWeghe06,Iliopoulos14}, facilitating the comparison between positions of objects at different times. Another intended extension of this work is to consider 3D poses as goal positions, this should also force the use of the three-dimensional QTC to represent more complex trajectories \citep{Mavridis14}, or the extension of $StarVars$ using quaternions to account for 3D spatial rotations.

This paper brings to the fore the possibility of guiding a sensory deprived autonomous agent using the perspectives of other autonomous agents modelled as qualitative relations. The methods introduced here could accommodate guided agents with or without known orientation, which suggests that this research provides a general tool for guided navigation tasks.

\section*{Acknowledgement(s)}

The authors would like to thank the three anonymous reviewers for the thoughtful comments and suggestions that helped us to improve the submitted paper.

We acknowledge financial support from CAPES, CNPq and FAPESP (grant 2012/04089-3).


\bibliographystyle{apalike}

\begin{thebibliography}{}

\bibitem[Bianchi et~al., 2018]{Bianchi:2018}
Bianchi, R.~A., Santos, P.~E., Silva, I.~J., Celiberto, Jr, L.~A., and Lopez
  De~Mantaras, R. (2018).
\newblock Heuristically accelerated reinforcement learning by means of
  case-based reasoning and transfer learning.
\newblock {\em J. Intell. Robotics Syst.}, 91(2):301--312.

\bibitem[Cadena et~al., 2016]{Cadena16tro-SLAMfuture}
Cadena, C., Carlone, L., Carrillo, H., Latif, Y., Scaramuzza, D., Neira, J.,
  Reid, I., and Leonard, J. (2016).
\newblock Past, present, and future of simultaneous localization and mapping:
  Towards the robust-perception age.
\newblock {\em {IEEE Transactions on Robotics}}, 32(6):1309–1332.

\bibitem[Cangelosi and Ogata, 2016]{Cangelosi2016}
Cangelosi, A. and Ogata, T. (2016).
\newblock {\em Speech and Language in Humanoid Robots}, pages 1--32.
\newblock Springer Netherlands, Dordrecht.

\bibitem[Chen et~al., 2015]{chen_cohn_liu_wang_ouyang_yu_2015}
Chen, J., Cohn, A.~G., Liu, D., Wang, S., Ouyang, J., and Yu, Q. (2015).
\newblock A survey of qualitative spatial representations.
\newblock {\em The Knowledge Engineering Review}, 30(1):106–136.

\bibitem[Cohn and Renz, 2007]{QSRKRHandbook07}
Cohn, A.~G. and Renz, J. (2007).
\newblock Qualitative spatial reasoning.
\newblock In van Harmelen, F., Lifschitz, V., and Porter, B., editors, {\em
  Handbook of Knowledge Representation}. Elsevier.

\bibitem[Colombo et~al., 2017]{COLOMBO2017605}
Colombo, D., Serino, S., Tuena, C., Pedroli, E., Dakanalis, A., Cipresso, P.,
  and Riva, G. (2017).
\newblock Egocentric and allocentric spatial reference frames in aging: A
  systematic review.
\newblock {\em Neuroscience \& Biobehavioral Reviews}, 80(Supplement C):605 --
  621.

\bibitem[Coradeschi and Saffiotti, 2003]{CoradeschiSaffiotti.ras03}
Coradeschi, S. and Saffiotti, A. (2003).
\newblock An introduction to the anchoring problem.
\newblock {\em Robotics and Autonomous Systems}, 43(2-3):85--96.
\newblock Special issue on perceptual anchoring. Online at
  http://www.aass.oru.se/Agora/RAS02/.

\bibitem[Dantzig, 1990]{Dantzig:1990}
Dantzig, G.~B. (1990).
\newblock Origins of the simplex method.
\newblock In Nash, S.~G., editor, {\em A History of Scientific Computing},
  pages 141--151. ACM, New York, NY, USA.

\bibitem[de~Weghe, 2004]{vanDerWeghe04}
de~Weghe, N.~V. (2004).
\newblock {\em Representing and Reasoning about Moving Objects: A Qualitative
  Approach}.
\newblock Phd thesis, Ghent University.

\bibitem[{DeSouza} and {Kak}, 2002]{desouza02}
{DeSouza}, G.~N. and {Kak}, A.~C. (2002).
\newblock Vision for mobile robot navigation: a survey.
\newblock {\em IEEE Transactions on Pattern Analysis and Machine Intelligence},
  24(2):237--267.

\bibitem[Dylla et~al., 2007]{cosy:R3-R4-sailaway-aisb}
Dylla, F., Frommberger, L., Wallgr{\"{u}}n, J.~O., Wolter, D., W{\"{o}}lfl, S.,
  and Nebel, B. (2007).
\newblock Sailaway: Formalizing navigation rules.
\newblock In {\em AISB'07 ARTIFICIAL AND AMBIENT INTELLIGENCE SYMPOSIUM ON
  SPATIAL REASONING AND COMMUNICATION}.

\bibitem[Ekstrom et~al., 2018]{ekstrom18}
Ekstrom, A.~D., Spiers, H.~J., Bohbot, V.~D., and Rosenbaum, R.~S. (2018).
\newblock {\em Human Spatial Navigation}.
\newblock Princeton University Press.

\bibitem[Epstein et~al., 2017]{Epstein17}
Epstein, R., Patai, E.~Z., Julian, J., and Spiers, H. (2017).
\newblock The cognitive map in humans: Spatial navigation and beyond.
\newblock {\em Nature Neuroscience}, 20:1504--1513.

\bibitem[Freksa, 1992]{Freksa92}
Freksa, C. (1992).
\newblock Using orientation information for qualitative spatial reasoning.
\newblock In Frank, A.~U., Campari, I., and Formentini, U., editors, {\em
  Theories and Methods of Spatio-Temporal Reasoning in Geographic Space}, pages
  162--178, Berlin, Heidelberg. Springer Berlin Heidelberg.

\bibitem[Frommberger et~al., 2007]{opra-composition}
Frommberger, L., Lee, J.~H., Wallgr{\"{u}}n, J.~O., and Dylla, F. (2007).
\newblock Composition in opram.
\newblock Technical Report 013-02/2007, SFB/TR 8 Spatial Cognition;
  http://www.sfbtr8.uni-bremen.de/.

\bibitem[Games and Howell, 1976]{games-howell}
Games, P.~A. and Howell, J.~F. (1976).
\newblock Pairwise multiple comparison procedures with unequal n's and/or
  variances: A monte carlo study.
\newblock {\em Journal of Educational Statistics}, 1(2):113--125.

\bibitem[Gribble et~al., 1998]{Gribble1998}
Gribble, W.~S., Browning, R.~L., Hewett, M., Remolina, E., and Kuipers, B.~J.
  (1998).
\newblock {\em Integrating vision and spatial reasoning for assistive
  navigation}, pages 179--193.
\newblock Springer Berlin Heidelberg, Berlin, Heidelberg.

\bibitem[Gupta, 2009]{rlinear}
Gupta, R.~K. (2009).
\newblock {\em Linear Programming}.
\newblock Krishna Prakashan.

\bibitem[Harnad, 1990]{HARNAD1990335}
Harnad, S. (1990).
\newblock The symbol grounding problem.
\newblock {\em Physica D: Nonlinear Phenomena}, 42(1):335 -- 346.

\bibitem[Homem et~al., 2017]{HomemPSBM17}
Homem, T. P.~D., Perico, D.~H., Santos, P.~E., da~Costa~Bianchi, R.~A., and
  de~M{\'{a}}ntaras, R.~L. (2017).
\newblock Retrieving and reusing qualitative cases: An application in
  humanoid-robot soccer.
\newblock {\em {AI} Commun.}, 30(3-4):251--265.

\bibitem[Iliopoulos et~al., 2014]{Iliopoulos14}
Iliopoulos, K., Bellotto, N., and Mavridis, N. (2014).
\newblock From sequence to trajectory and vice versa: Solving the inverse qtc
  problem and coping with real-world trajectories.
\newblock {\em AAAI Spring Symposium - Technical Report}, pages 57--64.

\bibitem[Klippel et~al., 2005]{KLIPPEL2005311}
Klippel, A., Tappe, H., Kulik, L., and Lee, P.~U. (2005).
\newblock Wayfinding choremes—a language for modeling conceptual route
  knowledge.
\newblock {\em Journal of Visual Languages \& Computing}, 16(4):311 -- 329.
\newblock Perception and ontologies in visual, virtual and geographic space.

\bibitem[Kreutzmann et~al., 2013]{Kreutzmann_Wolter_Dylla_Lee_2013}
Kreutzmann, A., Wolter, D., Dylla, F., and Lee, J.~H. (2013).
\newblock Towards safe navigation by formalizing navigation rules.
\newblock {\em TransNav, the International Journal on Marine Navigation and
  Safety of Sea Transportation}, 7(2):161--168.

\bibitem[Kruse et~al., 2013]{KRUSE20131726}
Kruse, T., Pandey, A.~K., Alami, R., and Kirsch, A. (2013).
\newblock Human-aware robot navigation: A survey.
\newblock {\em Robotics and Autonomous Systems}, 61(12):1726 -- 1743.

\bibitem[Lee, 2013]{JaeHeeLee}
Lee, J.~H. (2013).
\newblock {\em Qualitative Reasoning About Relative Directions}.
\newblock Doctorate in mathematics and informatics, University of Bremen,
  Bremen, Germany.

\bibitem[Lee et~al., 2013]{Lee13starvars}
Lee, J.~H., Renz, J., and Wolter, D. (2013).
\newblock Starvars -- effective reasoning about relative directions.
\newblock In {\em INTL. JOINT CONF. ON ARTIFICAL INTELLIGENCE (IJCAI)}, pages
  976--982.

\bibitem[Levesque and Lakemeyer, 2008]{levesque08}
Levesque, H. and Lakemeyer, G. (2008).
\newblock Chapter 23 cognitive robotics.
\newblock {\em Foundations of Artificial Intelligence}, 3.

\bibitem[Levitt and Lawton, 1990]{LEVITT1990305}
Levitt, T.~S. and Lawton, D.~T. (1990).
\newblock Qualitative navigation for mobile robots.
\newblock {\em Artificial Intelligence}, 44(3):305 -- 360.

\bibitem[Ligozat, 2013]{ligozat2013}
Ligozat, G.~E. (2013).
\newblock {\em Qualitative Spatial and Temporal Reasoning}.
\newblock ISTE. Wiley.

\bibitem[Liu and Zhang, 2019]{liu19}
Liu, R. and Zhang, X. (2019).
\newblock A review of methodologies for natural-language-facilitated humanoid
  robot cooperation.
\newblock {\em International Journal of Advanced Robotic Systems},
  16(3):1729881419851402.

\bibitem[L{\"u}cke et~al., 2011]{LueckeEtAL2011}
L{\"u}cke, D., Mossakowski, T., and Moratz, R. (2011).
\newblock Streets to the opra - finding your destination with imprecise
  knowledge.
\newblock In Renz, J., Cohn, A.~G., and W{\"o}lfl, S., editors, {\em IJCAI
  Workshop on Benchmarks and Applications of Spatial Reasoning}, pages 25--32.

\bibitem[Martin et~al., 2017]{martin17}
Martin, A.~P., Santos, P.~E., and Safi-Samghabadi, M. (2017).
\newblock Consistency check in a multiple viewpoint system for reasoning about
  occlusion.
\newblock In Benferhat, S., Tabia, K., and Ali, M., editors, {\em Advances in
  Artificial Intelligence: From Theory to Practice}, pages 523--532, Cham.
  Springer International Publishing.

\bibitem[Mavridis, 2007]{Mavridis07}
Mavridis, N. (2007).
\newblock {\em Grounded Situation Models for Situated Conversational
  Assistants}.
\newblock Phd thesis, Massachusetts Institute of Technology.

\bibitem[Mavridis, 2015]{Mavridis:2015}
Mavridis, N. (2015).
\newblock A review of verbal and non-verbal human-robot interactive
  communication.
\newblock {\em Robot. Auton. Syst.}, 63(P1):22--35.

\bibitem[Mavridis et~al., 2014]{Mavridis14}
Mavridis, N., Bellotto, N., Iliopoulos, K., and Van~de Weghe, N. (2014).
\newblock Qtc3d: Extending the qualitative trajectory calculus to three
  dimensions.
\newblock {\em Information Sciences}, 26.

\bibitem[McClelland et~al., 2016]{MCCLELLAND201673}
McClelland, M., Campbell, M., and Estlin, T. (2016).
\newblock Qualitative relational mapping and navigation for planetary rovers.
\newblock {\em Robotics and Autonomous Systems}, 83:73 -- 86.

\bibitem[McDonald, 2014]{mcdonald_welch_anova}
McDonald, J. (2014).
\newblock {\em Handbook of Biological Statistics}.
\newblock Sparky House Publishing, Baltimore, Maryland, 3 edition.

\bibitem[Moratz, 2006]{OPRAMoratz06}
Moratz, R. (2006).
\newblock Representing relative direction as a binary relation of oriented
  points.
\newblock In {\em ECAI 2006 Proceedings of the 17th European Conference on
  Artificial Intelligence}.

\bibitem[Moratz et~al., 2003]{Moratz03}
Moratz, R., Nebel, B., and Freksa, C. (2003).
\newblock Qualitative spatial reasoning about relative position.
\newblock In Freksa, C., Brauer, W., Habel, C., and Wender, K.~F., editors,
  {\em Spatial Cognition III}, pages 385--400, Berlin, Heidelberg. Springer
  Berlin Heidelberg.

\bibitem[Mossakowski and Moratz, 2012]{OPRA12}
Mossakowski, T. and Moratz, R. (2012).
\newblock Qualitative reasoning about relative direction of oriented points.
\newblock {\em Artificial Intelligence}, 180-181:34--45.

\bibitem[Mourikis and Roumeliotis, 2006]{mourikis}
Mourikis, A.~I. and Roumeliotis, S.~I. (2006).
\newblock Predicting the performance of cooperative simultaneous localization
  and mapping (c-slam).
\newblock {\em The International Journal of Robotics Research},
  25(12):1273--1286.

\bibitem[Mueggler et~al., 2014]{7017662}
Mueggler, E., Faessler, M., Fontana, F., and Scaramuzza, D. (2014).
\newblock Aerial-guided navigation of a ground robot among movable obstacles.
\newblock In {\em IEEE INTERNATIONAL SYMPOSIUM ON SAFETY, SECURITY, AND RESCUE
  ROBOTICS}, pages 1--8.

\bibitem[{Muthugala} and {Jayasekara}, 2018]{8298518}
{Muthugala}, M. A. V.~J. and {Jayasekara}, A. G. B.~P. (2018).
\newblock A review of service robots coping with uncertain information in
  natural language instructions.
\newblock {\em IEEE Access}, 6:12913--12928.

\bibitem[Nikolaidis et~al., 2018]{Nikolaidis:2018}
Nikolaidis, S., Kwon, M., Forlizzi, J., and Srinivasa, S. (2018).
\newblock Planning with verbal communication for human-robot collaboration.
\newblock {\em ACM Trans. Hum.-Robot Interact.}, 7(3):22:1--22:21.

\bibitem[Pereira et~al., 2013]{DBLP:conf/bracis/PereiraCSM13}
Pereira, V.~F., Cozman, F.~G., Santos, P.~E., and Martins, M.~F. (2013).
\newblock A qualitative-probabilistic approach to autonomous mobile robot self
  localisation and self vision calibration.
\newblock In {\em BRAZILIAN CONFERENCE ON INTELLIGENT SYSTEMS (BRACIS)}, pages
  157--162.

\bibitem[Perico et~al., 2018]{perico2018humanoid}
Perico, H.~D., Homem, P. D.~T., Almeida, C.~A., Silva, J.~I., Vilão, O.~C.,
  Ferreira, N.~V., and Bianchi, A. C.~R. (2018).
\newblock Humanoid robot framework for research on cognitive robotics.
\newblock {\em Journal of Control, Automation and Electrical Systems}, pages
  470--479.

\bibitem[Renz and Mitra, 2004]{STARRenz04}
Renz, J. and Mitra, D. (2004).
\newblock Qualitative direction calculi with arbitrary granularity.
\newblock In {\em PACIFIC RIM INTERNATIONAL CONFERENCE ON ARTIFICIAL
  INTELLIGENCE (PRICAI)}, pages 65--74. Springer.

\bibitem[{RoboCup Federation}, 2020]{RoboCup}
{RoboCup Federation} (2020).
\newblock Robocup.org.

\bibitem[Santos et~al., 2016a]{santos15}
Santos, P., Ligozat, G., and Safi-Samghabad, M. (2016a).
\newblock An occlusion calculus based on an interval algebra.
\newblock In {\em Proceedings - 2015 Brazilian Conference on Intelligent
  Systems, BRACIS 2015}, Proceedings - 2015 Brazilian Conference on Intelligent
  Systems, BRACIS 2015, pages 128--133. Institute of Electrical and Electronics
  Engineers Inc.
\newblock 4th Brazilian Conference on Intelligent Systems, BRACIS 2015 ;
  Conference date: 04-11-2015 Through 07-11-2015.

\bibitem[Santos et~al., 2009]{Santos09}
Santos, P.~E., Dee, H., and Fenelon, V. (2009).
\newblock Qualitative robot localisation using information from cast shadows.
\newblock In {\em IEEE INTERNATIONAL CONFERENCE ON ROBOTICS AND AUTOMATION
  (ICRA)}, pages 220--225.

\bibitem[Santos et~al., 2016b]{valquiria16}
Santos, P.~E., Martins, M.~F., Fenelon, V., Cozman, F.~G., and Dee, H.~M.
  (2016b).
\newblock Probabilistic self-localisation on a qualitative map based on
  occlusions.
\newblock {\em Journal of Experimental \& Theoretical Artificial Intelligence},
  28(5):781--799.

\bibitem[{Suárez Bonilla} and {Ruiz Ugalde}, 2019]{8675641}
{Suárez Bonilla}, F. and {Ruiz Ugalde}, F. (2019).
\newblock Automatic translation of spanish natural language commands to control
  robot comands based on lstm neural network.
\newblock In {\em 2019 Third IEEE International Conference on Robotic Computing
  (IRC)}, pages 125--131.

\bibitem[Tellex et~al., 2011]{Tellex_2011}
Tellex, S., Kollar, T., Dickerson, S., Walter, M.~R., Banerjee, A.~G., Teller,
  S., and Roy, N. (2011).
\newblock Approaching the symbol grounding problem with probabilistic graphical
  models.
\newblock {\em AI Magazine}, 32(4):64--76.

\bibitem[Tellex and Roy, 2006]{Tellex:2006}
Tellex, S. and Roy, D. (2006).
\newblock Spatial routines for a simulated speech-controlled vehicle.
\newblock In {\em Proceedings of the 1st ACM SIGCHI/SIGART Conference on
  Human-robot Interaction}, HRI '06, pages 156--163, New York, NY, USA. ACM.

\bibitem[Thrun et~al., 2005]{Thrun:2005}
Thrun, S., Burgard, W., and Fox, D. (2005).
\newblock {\em Probabilistic Robotics}.
\newblock The MIT Press.

\bibitem[Tversky, 2014]{Tversky2014}
Tversky, B. (2014).
\newblock {\em Visualizing Thought}, pages 3--40.
\newblock Springer New York, New York, NY.

\bibitem[Tversky and Lee, 1999]{tversky99}
Tversky, B. and Lee, P.~U. (1999).
\newblock Pictorial and verbal tools for conveying routes.
\newblock In Freksa, C. and Mark, D.~M., editors, {\em Spatial Information
  Theory. Cognitive and Computational Foundations of Geographic Information
  Science}, pages 51--64, Berlin, Heidelberg. Springer Berlin Heidelberg.

\bibitem[Van~de Weghe et~al., 2006]{vanDerWeghe06}
Van~de Weghe, N., Cohn, A., Tré, G., and De~Maeyer, P. (2006).
\newblock A qualitative trajectory calculus as a basis for representing moving
  objects in geographical information systems.
\newblock {\em CONTROL AND CYBERNETICS}, 35:97--119.

\bibitem[Wallgr\"un, 2010]{wall10}
Wallgr\"un, J.~O. (2010).
\newblock Qualitative spatial reasoning for topological map learning.
\newblock {\em Spatial Cognition \& Computation}, 10(4):207--246.

\bibitem[Welch, 1951]{welch}
Welch, B.~L. (1951).
\newblock On the comparison of several mean values: An alternative approach.
\newblock {\em Biometrika}, 38(3-4):330--336.

\bibitem[Wolter et~al., 2008]{Wolter2008}
Wolter, D., Freksa, C., and Jan~Latecki, L. (2008).
\newblock {\em Towards a Generalization of Self-localization}, pages 105--134.
\newblock Springer Berlin Heidelberg, Berlin, Heidelberg.

\bibitem[{Zhang} et~al., 2019]{8778650}
{Zhang}, S., {Jiang}, J., {He}, Z., {Zhao}, X., and {Fang}, J. (2019).
\newblock A novel slot-gated model combined with a key verb context feature for
  task request understanding by service robots.
\newblock {\em IEEE Access}, 7:105937--105947.

\end{thebibliography}

\end{document}